\documentclass{interact}
\usepackage{epstopdf}
\usepackage[caption=false]{subfig}

\usepackage{amsmath,amssymb,amsthm}
\usepackage{amsfonts}
\usepackage{bm}
\usepackage{booktabs}
\usepackage[english]{babel}
\usepackage{subfig}
\usepackage{graphicx}
\usepackage{subcaption}
\usepackage[colorlinks=true, allcolors=blue]{hyperref}
\usepackage{indentfirst}
\usepackage{multirow}
\usepackage{appendix}
\usepackage{algorithm}
\usepackage{algpseudocode}
\usepackage[table]{xcolor}

\theoremstyle{plain}

\theoremstyle{definition}
\newtheorem{theorem}{Theorem}

\theoremstyle{remark}

\begin{document}
\title{LFR-PINO: A Layered Fourier Reduced Physics-Informed Neural Operator for Parametric PDEs}

\author{
 \name{Jing Wang\textsuperscript{a} 
 and Biao Chen\textsuperscript{b} 
 and Hairun Xie\textsuperscript{c} 
 and Rui Wang\textsuperscript{a}
 and Yifan Xia\textsuperscript{b} 
 and Jifa Zhang\textsuperscript{b}\thanks{CONTACT A.~N. Author. Email: jfzhang@zju.edu.cn}
 and Hui Xu\textsuperscript{a} \thanks{CONTACT A.~N. Author. Email: dr.hxu@sjtu.edu.cn}}
\affil{\textsuperscript{a}School of Aeronautics and Astronautics, Shanghai Jiao Tong University, Shanghai, China\\
 \textsuperscript{b}School of Aeronautics and Astronautics, Zhejiang University, Hangzhou, China\\
 \textsuperscript{c}Key Laboratory for Satellite Digitalization Technology, Innovation Academy for Microsatellites of CAS, Shanghai, China}
 }


\maketitle

\begin{abstract}
Physics-informed neural operators have emerged as a powerful paradigm for solving parametric partial differential equations (PDEs), particularly in the aerospace field, enabling the learning of solution operators that generalize across parameter spaces. However, existing methods either suffer from limited expressiveness due to fixed basis/coefficient designs, or face computational challenges due to the high dimensionality of the parameter-to-weight mapping space.
We present LFR-PINO, a novel physics-informed neural operator that introduces two key innovations: (1) a layered hypernetwork architecture that enables specialized parameter generation for each network layer, and (2) a frequency-domain reduction strategy that significantly reduces parameter count while preserving essential spectral features. This design enables efficient learning of a universal PDE solver through pre-training, capable of directly handling new equations while allowing optional fine-tuning for enhanced precision.
The effectiveness of this approach is demonstrated through comprehensive experiments on four representative PDE problems, where LFR-PINO achieves 22.8\%-68.7\% error reduction compared to state-of-the-art baselines. Notably, frequency-domain reduction strategy reduces memory usage by 28.6\%-69.3\% compared to Hyper-PINNs while maintaining solution accuracy, striking an optimal balance between computational efficiency and solution fidelity.
\end{abstract}


\section{Introduction}

Parametric partial differential equations (PDEs) play a crucial role in modern scientific computing and engineering applications, serving as the cornerstone for modeling complex physical phenomena across diverse fields. From optimizing aerodynamic designs in aerospace engineering~\cite{li2022machine} to enhancing structural safety in civil engineering~\cite{han2018solving}, these equations are essential for understanding and predicting system behaviors under varying conditions. 
The ability to efficiently solve parametric PDEs has become increasingly critical as industries demand faster, more accurate solutions for real-time decision-making and optimization.

Traditional numerical methods, including Finite Element Methods (FEM)~\cite{ZIENKIEWICZ20131}, Finite Difference Methods (FDM)~\cite{godunov1959finite}, and Finite Volume Methods (FVM)~\cite{eymard2000finite}, while well-established, face significant challenges in the context of parametric PDEs. These methods require extensive domain discretization and repeated solving for different parameter values, leading to computational costs that scale poorly with problem complexity.

The emergence of machine learning approaches has opened new avenues for addressing these challenges. \textbf{Data-driven neural operators}, such as DeepONet~\cite{lu2021learning} and Fourier Neural Operator (FNO)~\cite{li2020fourier}, have demonstrated remarkable success in learning continuous mappings between function spaces. These methods can reduce solution time from hours to seconds once trained. 
However, these methods often demand extensive high-quality datasets, and their generalization capability to out-of-distribution cases or complex geometries remains a challenge.

Conversely, \textbf{physics-driven neural operators}, such as PI-DeepONet~\cite{wang_learning_2021} and Meta-Auto-Decoder (MAD)~\cite{MADHuang2021}, integrate governing physical laws directly into the learning process, providing the advantage of unsupervised adaptation to new parameters. Despite this, balancing computational efficiency with representation accuracy, especially for high-dimensional and multi-scale problems, poses a significant challenge. Extensions like Hyper-PINNs~\cite{belbute-peres_hyperpinn_nodate} attempt to map variable parameters to network weights via hypernetworks, yet the resultant high-dimensional weight space incurs severe computational overhead.

In this study, we propose the Layered Fourier-Reduced Physics-Informed Neural Operator (LFR-PINO) to address fundamental challenges for solving parametric PDEs. First, to tackle the challenge of excessive memory consumption while maintaining representation ability, we introduce a Fourier-reduced framework inspired by the frequency principle~\cite{xu2019training, xu2018Fourier}. This framework projects network weights into the frequency domain and retains only the most significant modes, reducing memory requirements by up to 70\% while preserving solution fidelity. Second, to address the challenge of capturing multi-scale phenomena efficiently, we design a layered architecture that explicitly models different frequency bands, achieving accurate representation with fewer parameters. 
Finally, we use physics-informed constraints that enable the model to learn directly from governing equations without requiring extensive datasets.

In summary, our contributions are as follows: 
(1) We propose a unified framework that categorizes existing methods according to their spectral decomposition strategies, and introduce Neural Operators with Dynamic Basis and Coefficients as a novel perspective;
(2) We develop the Layered Fourier Reduced Framework, which synergistically combines Fourier reparameterization and layered hypernetworks to address the computational challenges of parametric PDEs;
(3) We provide both theoretical analyses and experimental validations that demonstrate the improved representation capability enhanced generalization and optimization stability of our approach.

\section{Related Works}

\paragraph{Physics-informed Neural Networks}
With the rise of artificial intelligence~\cite{lecun2015deep}, neural networks (NNs) have become a powerful tool for solving PDEs through unsupervised learning. Methods such as Physics-informed Neural Networks (PINNs)~\cite{RAISSI2019686}, Deep Galerkin Method (DGM)~\cite{sirignano2018dgm}, Deep Ritz Method (DRM)~\cite{yu2018deep}, and PhyCRNet~\cite{rao2023encoding} incorporate governing equations and boundary conditions into the loss function or network architecture to approximate PDE solutions. While these approaches have shown success in applications like turbulence modeling and complex flow simulations~\cite{raissi2020,xu2021explore}, they are inherently limited to solving single PDE instances, making them impractical for many-query scenarios involving varied PDE parameters.

\paragraph{Data-driven Neural Operators}
To address the limitations of traditional methods in solving parametric PDEs, data-driven neural operators have emerged as a novel approach, enabling the mapping of input function spaces to output solution spaces. Various architectures have been proposed to learn such mappings: DeepONet~\cite{lu2021learning} and Fourier Neural Operator (FNO)~\cite{li2020fourier} represent two fundamental approaches, while PDE-Net~\cite{long2018pde} combines differential operations with convolutional neural networks to discover and approximate underlying PDEs. Graph-based methods, such as Graph Neural Operator (GNO)~\cite{li2020neural}, have demonstrated effectiveness in handling irregular domains and complex geometries. More recently, transformer-based architectures~\cite{deng2023prediction} have been explored to capture complex correlations in the input space through self-attention mechanisms.
These neural operator models excel at generalizing to new PDE parameters without retraining, offering a promising alternative for high-dimensional and complex parametric PDEs. However, their reliance on substantial labeled data during training poses challenges in scenarios where labeled data are scarce, and their generalization capability may degrade when applied to out-of-distribution cases.

\paragraph{Physics-informed Neural Operators}
Physics-informed neural operators combine neural operator frameworks with physics-based constraints, enabling solution mapping for parametric PDEs without extensive labeled data. Many attempts focused on meta-learning algorithms (MAML, Reptile)~\cite{penwarden_metalearning_2023, liu_novel_2022, guo__metapinns_2024} to optimize PINNs initialization, or employed multi-task learning strategies~\cite{zou2023hydra} to adapt to different PDEs. However, these methods primarily rely on learning to adapt to different PDEs rather than establishing explicit mappings, necessitating additional training for each new PDE task.
Recent advances include PI-DeepONet~\cite{wang_learning_2021}, which incorporates physical laws into DeepONet's training process, and Meta-Auto-Decoder (MAD)~\cite{MADHuang2021}, which constructs nonlinear trial manifolds for solution spaces. Hyper-PINNs~\cite{belbute-peres_hyperpinn_nodate} employs a hyper-network~\cite{ha2016hypernetworks} to parameterize PINN weights as functions of PDE parameters, enabling a compact representation of multiple solutions. Despite these innovations, challenges remain in achieving computational efficiency and solution accuracy, motivating our current work.

\section{Methodology}

We consider a general framework for parametric PDEs that encompasses variations in physical parameters, boundary conditions, and initial conditions. These equations are formulated in an operator form with a finite time horizon $T\in \mathbb{R}^+$ and a bounded domain $\Omega \subset \mathbb{R}^d$:
\begin{equation}
\begin{cases}
\mathcal{R}_{x}^{\eta_{\mathrm{r}}}u=0, &  x \in [0,T]\times \Omega, \\
\mathcal{B}_{x}^{\eta_{\mathrm{bc}}}u=0, &  x \in  [0,T]\times \partial \Omega, \\
\mathcal{I}_{x}^{\eta_{\mathrm{ic}}}u=0, & x \in \{0\}\times \Omega,
\end{cases}  
\label{pde}
\end{equation}
where $\mathcal{R}$, $\mathcal{B}$ and $\mathcal{I}$ denote the differential operators, boundary conditions, and initial conditions parameterized by $\eta_{\mathrm{r}}$, $\eta_{\mathrm{bc}}$ and $\eta_{\mathrm{ic}}$, respectively. $x \in \mathbb{R}^{d_x}$ denotes the independent variable in spatiotemporal dependent PDEs.
$u: \mathbb{R}^{d_x} \longmapsto \mathbb{R}^{d_u}$ denotes the solution of PDEs, mapping from the spatiotemporal domain to a vector space.

The fundamental objective in solving parametric PDEs is to approximate a deterministic mapping that maps each parameter set $\eta=\{\eta_{\mathrm{r}}, \eta_{\mathrm{bc}}, \eta_{\mathrm{ic}}\}$, which can include but is not limited to the differential parameter, the boundary condition parameter, or the initial condition parameter, to its corresponding solution $u$ in such a way that the solution space is well-behaved with respect to variations in the parameters.
Given a parameter space $\mathcal{H}$ and a solution space $\mathcal{U}=\mathcal{U}(\Omega;\mathbb{R}^{d_u})$, both of which are Banach spaces, we define an infinite-dimensional neural operator to map any parameter $\eta$ to its corresponding solution $u_{\eta}$:
\begin{equation}
    G_{\infty}: \mathcal{H} \longmapsto \mathcal{U}, \eta \longmapsto u_{\eta},
\end{equation}
where $\eta \in \mathcal{H}$ is the parameter vector, which involves the differential parameter $\eta_{\mathrm{r}}$, the boundary condition parameter $\eta_{\mathrm{bc}}$, and the initial condition parameter $\eta_{\mathrm{ic}}$ in this study.

The space $\mathcal{H}$ and $\mathcal{U}$ are both infinite-dimensional and it is hard to model the precise space. 
Instead of attempting to directly approximate $G_{\infty}$, one can assume that there exists a basis of a high-fidelity discretization of $\mathcal{H}$ and $\mathcal{U}$ and try to exploit possible finite-dimensional approximation for a finite set of discrete sets $\{\eta^j, u_{\eta}^j\}_{j=1}^N$. 

Given a finite-dimensional parameter space $\mathcal{H_N}$ and a finite-dimensional solution space $\mathcal{S_M}$, 
representing finite-dimensional subspaces of $\mathcal{H}$ and $\mathcal{S}$, respectively, the neural operator $G$ is formulated as a discrete approximation of  $G_\infty$, represented as follows:
\begin{equation}
G: \mathcal{H}_N \longmapsto \mathcal{U}_M, \eta \longmapsto u_{\eta},
\end{equation}
such that the solution set can be approximated by a compact subset:
\begin{equation}
G(\mathcal{H}_N) = \{ G(\eta) | \eta \in \mathcal{H}_N \} \subset \mathcal{U}_M.
\end{equation}

\subsection{Spectral Perspective for Neural Operator}
Before presenting our approach, we first establish a theoretical foundation that bridges classical spectral methods with modern neural network architectures. This unified perspective not only provides insights into existing methods but also motivates our proposed framework.

According to the generalized spectral methods in numerical analysis, the weak solution of $u$ is traditionally approximated as a linear combination of spectral coefficients $\{\alpha_n\}_{n=1}^N$ and their associated spectral basis $\{\phi_n\}_{n=1}^N$ as follows: 
\begin{equation}
    u \approx \sum_{n=1}^N{\alpha_n\phi_n}.
    \label{eq:basis}
\end{equation}
Theoretically, $\phi_n$ can be constructed by proper orthogonal decomposition eigenmodes, Fourier modes, orthogonal function spaces, operator normal modes, or Koopman modes, etc.  
While coefficients can be obtained via methods like the Legendre spectral element method or the Fourier spectral element method, both are extensively utilized in the domain.

Given the exceptional ability of neural networks to capture intricate nonlinear relationships and patterns, recent years have witnessed the emergence of various neural network-based approaches for solving parametric PDEs. 
From the spectral representation perspective, existing neural network-based approaches can be systematically categorized into three classes based on their treatment of basis and coefficients:
\paragraph{1. Neural Operator with Dynamic Coefficients}
 This class maintains fixed basis functions for all the PDEs while employing neural networks to learn parameter-dependent coefficients. The solution is approximated as: \begin{equation} u_{\eta}(x) \approx \sum_{n=1}^N \alpha_n(\eta) \phi_n(x). \end{equation}
A representative example is DeepONet, which utilizes two separate networks: one for encoding parameters $\eta$ to generate coefficients $\alpha_n(\eta)$, and another for producing parameter-independent basis functions $\phi_n(x)$. The solution is approximated as $ u_{\eta}(x) \approx \sum_{n=1}^N \alpha_n^{NN}(\eta;\theta_\alpha) \phi_n^{NN}(x;\theta_\phi)$.

\paragraph{2. Neural Operator with Dynamic Basis}
 This category employs neural networks to construct parameter-dependent basis functions while maintaining fixed coefficients for all the PDEs. 
The solution is approximated as: \begin{equation} u_{\eta}(x) \approx \sum_{n=1}^N \alpha_n \phi_n(\eta, x). \end{equation}
Both FNO and MAD fall into this class, though they differ in their basis construction mechanisms. In FNO, the basis functions are computed through a series of Fourier layers and spectral convolutions, where the outputs of all channels in the penultimate layer serve as the basis functions, and the weights in the final linear layer act as the fixed coefficients. The solution is approximated as  $u_{\eta}(x) \approx \sum_{n=1}^N \alpha_n \cdot \phi_n^{FNO}( \eta;x,\theta_\phi).$ 
Similarly, leveraging the weights in the last layer as the coefficients $\alpha_n$ and considering the outputs of the preceding neurons as the basis functions $\phi_n$, the MAD approach can be formulated as
    $u_{\eta}(x) \approx \sum_{n=1}^N \alpha_n \cdot \phi_n^{NN}(x, \eta;\theta_\phi).$

\paragraph{3. Neural Operator with Dynamic Basis and Coefficients}
 The most flexible approach involves simultaneously learning both parameter-dependent basis functions and coefficients.  The solution takes the form: 
\begin{equation} 
u_{\eta}(x) \approx \sum_{n=1}^N \alpha_n(\eta) \cdot \phi_n(\eta, x). 
\end{equation}
Methods like Hyper-PINNs~\cite{belbute-peres_hyperpinn_nodate} exemplify this category, where a hypernetwork is employed to generate the parameters of the primary network based on the input parameter $\eta$. By leveraging the weights in the last layer of the primary network as coefficients and considering the outputs of the preceding neurons as basis functions, the Hyper-PINNs can be formulated as $u_{\eta}(x) \approx \sum_{n=1}^N \alpha_n^{NN}(\eta;\theta_\alpha) \cdot \phi_n^{NN}(x;\theta_\phi(\eta)),$ where $\theta_\phi(\eta)$ represents the parameter-dependent weights generated by the hypernetwork.

The above three categories of methods represent a fundamental trade-off in neural operator design. While DeepONet achieves stability through fixed basis functions, it may lack expressiveness for problems with varying solution structures. FNO and MAD provide more flexibility through adaptive basis functions but may face challenges in coefficient optimization. Hyper-PINNs theoretically offer the most flexibility by learning both components. However, their current implementation faces two main challenges:

\begin{itemize}
    \item \textbf{Computational Complexity:} The hypernetwork generates all parameters of the main network, resulting in $\mathcal{O}(n^2)$ parameters where $n$ is the network width. This quadratic scaling becomes prohibitive for large-scale problems.    
    \item \textbf{Optimization Instability:} The simultaneous optimization of both basis functions and coefficients creates a highly coupled system. This coupling leads to ill-conditioned optimization landscapes and potential gradient interference between the hypernetwork and main network.
\end{itemize}

In the following, we aim to address the limitations of Hyper-PINNs and leverage the advantages of dynamic basis and coefficients.

\subsection{Layered Fourier Reduced Physics-informed Neural Operator}

\begin{figure}[htbp]
\centering
\includegraphics[width=\textwidth]{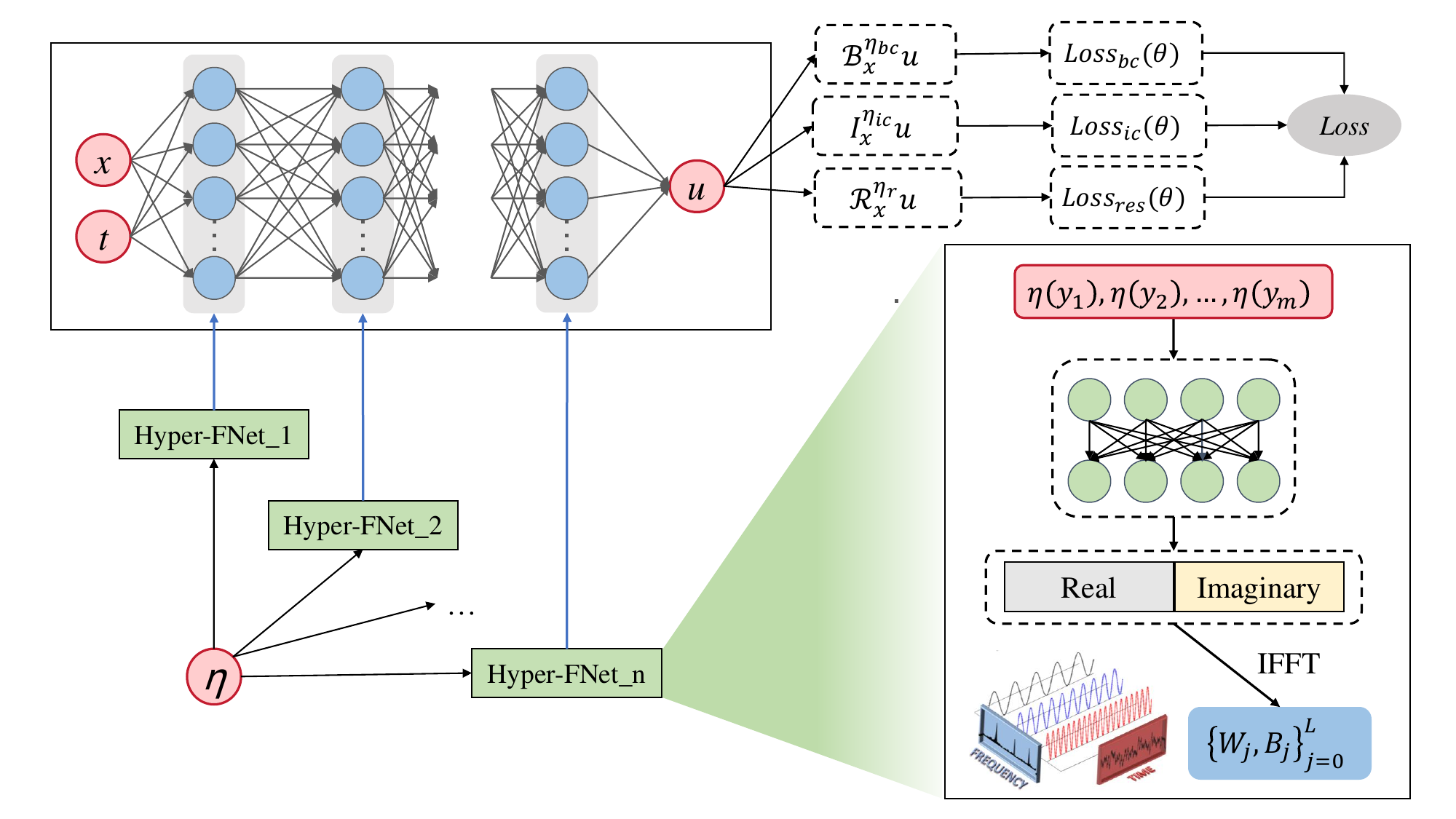}
    \caption{Architecture of the LFR-PINO framework}
    \label{fig:model}
\end{figure}

To achieve both computational efficiency and expressive power in neural operators, we propose the Layered Fourier Reduced Physics-informed Neural Operator (LFR-PINO), as illustrated in Figure~\ref{fig:model}. This novel architecture integrates hypernetworks with layered Fourier reduction technology, enabling efficient adaptation to diverse partial differential equations through unsupervised learning. 

\subsubsection{Network Architecture}

The LFR-PINO architecture consists of two key components: a main network for solution approximation and a set of layered Fourier reduced hyper networks for efficient weight generation of the main network.

\paragraph{Main Net}
The Main Net is an \(L\)-layer deep neural network mapping spatiotemporal coordinates \(x\) to PDE solutions across varying parameters \(\eta\):
\begin{equation}
\begin{aligned}
\gamma^{(0)} & = x, \\
\gamma^{(l)}(x, \eta) & = \sigma\left(W^{(l)}(\eta) \gamma^{(l-1)}(x, \eta)+b^{(l)}(\eta)\right), \quad l=1,2, \cdots, L-1, \\
f(x, \eta) & = W^{(L)}(\eta) \gamma^{(L)}(x, \eta),
\end{aligned}
\label{eq:main_net}
\end{equation}
where \(\gamma^{(l)}\) is the \(l\)-th layer's output, \(\sigma(\cdot)\) is the activation function, and \(\{W^{(l)}(\eta), b^{(l)}(\eta)\}\) are parameter-dependent weights and biases. The parameter \(\eta\) is discretized as \(\eta=[\eta(y_1), \eta(y_2), \cdots, \eta(y_m)]\) with \(m\) as the number of sampling points.

\paragraph{Layered Fourier Reduced Hyper Net} 
Traditional hypernetworks map parameters $\eta$ directly to all network weights $\{W^{(i)}, b^{(i)}\}_{i=1}^L$ (collectively denoted as $\{W^{(i)}\}_{i=1}^L$ for brevity). To reduce computational complexity while maintaining expressiveness, we propose a two-stage approach combining layered architecture with frequency domain reduction.

First, we decompose the weight generation process across layers. For the $i$-th layer, a dedicated hypernetwork $F_i$ generates only that layer's weights:
\begin{equation}
     F_i: \mathcal{H}_N \longmapsto \mathcal{W}^{(i)}, \quad i=1,2,\cdots,L 
\end{equation}
where $\mathcal{H}_N$ is the parameter space and $\mathcal{W}^{(i)}$ is the weight space of the $i$-th layer.

To further enhance efficiency, instead of directly generating the high-dimensional weights, each hypernetwork outputs a reduced set of Fourier coefficients:
\begin{equation}
     F_i: \eta \mapsto \hat{W}^{(i)} = [\hat{W}^{(i)}_1(\eta), \hat{W}^{(i)}_2(\eta), ..., \hat{W}^{(i)}_p(\eta)],
\end{equation}
where $\hat{W}^{(i)}_k$ represents the $k$-th Fourier coefficient for the $i$-th layer's weights. These coefficients are obtained through the Fourier transform of the weight matrix:
\begin{equation}
\hat{W}^{(i)}_k = \sum_{j=1}^{N} W^{(i)}_j e^{-2\pi i \frac{kj}{N}},
\end{equation}
where $N$ denotes the total number of weights in layer $i$, and $p \ll N$ is the number of retained frequency components.

This frequency-domain representation offers a natural way to reduce dimensionality while preserving essential information. While the full Fourier basis would require $N$ components for perfect reconstruction, we retain only the first $p$ low-frequency modes, as they typically capture the most important weight patterns while high-frequency components often correspond to noise.

The actual weights are then reconstructed through the inverse Fourier transform:
\begin{equation}
{W}^{(i)}=\frac{1}{N} \sum_{k=0}^{p-1} \hat{W}^{(i)}_k e^{2\pi i \frac{kn}{N}}.
\end{equation}

\paragraph{Implementation Details}
Since Fourier coefficients $\hat{W}^{(i)}_k$ are complex-valued, our hypernetworks output real and imaginary parts separately, which are then combined to form the complete complex representation. The number of retained frequency components $p$ is chosen to balance the trade-off between dimensionality reduction and approximation accuracy. This ensures that the reduced representation maintains sufficient expressiveness to capture the necessary nonlinear relationships in the solution space.

\subsubsection{Two-phase Unsupervised Training Strategy}    
Our framework employs an unsupervised, two-phase training approach using physics-informed loss functions. 

For any PDE parameter $\eta\in \mathcal{H}_N$, we define the physics-informed loss $L_{\eta}$ as:
\begin{equation}
L_{\eta}[u]= \left|\left|\mathcal{R}_{x}^{\eta_{\mathrm{r}}}u\right|\right|^2_{L^2(0,T;L^2(\Omega))} + \lambda_{bc} \left|\left|\mathcal{B}_{x}^{\eta_{\mathrm{bc}}}u\right|\right|^2_{L^2(0,T;L^2(\partial\Omega))} + \lambda_{ic} \left|\left|\mathcal{I}_{x}^{\eta_{\mathrm{ic}}}u\right|\right|^2_{L^2(L^2(\Omega))},
\end{equation}
where $\lambda_{bc}$ and $\lambda_{ic}$ balance the PDE residual, boundary, and initial conditions. Using Monte Carlo sampling with points $\{x_j^r\}_{j=1}^{M_r}$, $\{x_j^{bc}\}_{j=1}^{M_{bc}}$, and $\{x_j^{ic}\}_{j=1}^{M_{ic}}$, we approximate this loss as:
\begin{equation}
\hat{L}_{\eta}[u] = \frac{1}{M_r} \sum_{i=1}^{M_{r}}\left|\left|\mathcal{R}_{x}^{\eta_{\mathrm{r}}}u(x_j^r)\right|\right|^2_2 +  \frac{\lambda_{bc}}{M_{bc}} \sum_{i=1}^{M_{bc}}\left|\left|\mathcal{B}_{x}^{\eta_{\mathrm{bc}}}u(x_j^{bc})\right|\right|^2_2 + \frac{\lambda_{ic}}{M_{ic}} \sum_{i=1}^{M_{ic}}\left|\left|\mathcal{I}_{x}^{\eta_{\mathrm{ic}}}u(x_j^{ic})\right|\right|^2_2.
\end{equation}

The training consists of two phases:

\paragraph{Pre-training Stage}
During pre-training, we aim to learn a universal solver capable of handling a broad class of PDEs. Given hypernetwork parameters $\theta=\{\theta_1, \dots, \theta_L\}$ and $M$ diverse PDE parameters sampled from different equation types, we solve:
\begin{equation}
\theta^*=\underset{\theta}{\arg \min } \sum_{i=1}^M \hat{L}_{\eta}[u_{\eta}(x;\theta)].
\end{equation}
This phase learns a meta-knowledge mapping from PDE parameters to network weights, effectively creating a universal approximator that can generalize across different PDE families. The diversity of training PDEs is crucial here, as it enables the network to capture common mathematical structures and solution patterns shared across equations. In many cases, the pre-trained model already achieves satisfactory performance on new PDEs without additional tuning.

\paragraph{Fine-tuning} 
While the pre-trained model often provides sufficient accuracy, fine-tuning can be employed when higher precision is desired. For new parameters $\eta_{new}$ representing previously unseen equations, we can leverage the universal knowledge learned during pre-training and adjust network parameters to minimize:
\begin{equation}
\theta^*_{new}=\underset{\theta}{\arg \min } \hat{L}_{\eta}^{new}[u_{\eta}(x;\theta)].
\end{equation}
This full fine-tuning strategy provides maximum flexibility for adapting the universal solver to specific PDEs. Alternative fine-tuning strategies include adjusting only the coefficients ($\theta_L$) or only the basis ($\{\theta_1, \dots, \theta_{L-1}\}$), which may be computationally advantageous for specific applications. In our study, the full fine-tuning is adopted to demonstrate the method's complete adaptation capability. Algorithm~\ref{alg:training} details the complete training procedure.

\begin{algorithm}[htbp]
\caption{Training Algorithm for LFR-PINO}
\label{alg:training}
\begin{algorithmic}[1]
\Require 
    \State Training parameters $\{\eta_i\}_{i=1}^M$, new parameter $\eta_{\text{new}}$
    \State Main network $f(x, \eta)$, hypernetworks $\{F_i\}_{i=1}^L$
    \State Physics-informed loss $\hat{L}_{\eta}[u]$
\Ensure Optimized parameters $\theta_{\text{new}}^*$

\Statex \textbf{Pre-training Stage}
\State Initialize hypernetwork parameters $\theta = \{\theta_i\}_{i=1}^L$
\For{epoch = 1 to Max\_Epochs}
    \For{$\eta$ in $\{\eta_i\}_{i=1}^M$}
        \For{$i = 1$ to $L$}
            \State Generate Fourier coefficients: $\hat{W}^{(i)} = F_i(\eta; \theta_i)$ \Comment{$\hat{W}^{(i)} \in \mathbb{C}^p, p \ll N$}
            \State Reconstruct weights: $W^{(i)} = \text{IFFT}(\hat{W}^{(i)})$ \Comment{$W^{(i)} \in \mathbb{R}^N$}
        \EndFor
        \State Compute solution: $u_{\eta}(x) = f(x, \eta; \{W^{(i)}\}_{i=1}^L)$
        \State Update $\theta$ by minimizing $\hat{L}_{\eta}[u_{\eta}]$
    \EndFor
\EndFor

\Statex \textbf{Fine-tuning Stage (Optional)}
\For{$i = 1$ to $L$}
    \State $\hat{W}^{(i)} = F_i(\eta_{\text{new}}; \theta_i^*)$, $W^{(i)} = \text{IFT}(\hat{W}^{(i)})$
\EndFor
\State Initialize $\theta_{\text{new}} = \{W^{(i)}\}_{i=1}^L$
\For{epoch = 1 to Max\_Epochs}
    \State Compute solution: $u_{\eta_{\text{new}}}(x) = f(x, \eta_{\text{new}}; \theta_{\text{new}})$
    \State Update $\theta_{\text{new}}$ by minimizing $\hat{L}_{\eta_{\text{new}}}[u_{\eta_{\text{new}}}]$
\EndFor
\State \Return $\theta_{\text{new}}^*$
\end{algorithmic}
\end{algorithm}

\subsection{Theoretical Analysis}

We establish the effectiveness of LFR-PINO through theoretical results addressing approximation capability and optimization stability.

\subsubsection{Approximation Guarantee}
\begin{theorem}[Weight Approximation]
For any weight matrix \(\mathbf{W}\) and error threshold \(\epsilon \geq 0\), there exists a reduced frequency-space representation \(\mathbf{W}_f\) such that:
\begin{equation}
    \|\mathbf{W} - \mathbf{W}_f\|_2 \leq \epsilon,
\end{equation}
where \(\|\cdot\|_2\) denotes the spectral norm.
\end{theorem}
This theorem guarantees that our frequency-domain reduction preserves the network's representational capacity. The proof follows from matrix approximation theory, where truncating high-frequency components yields a controlled approximation error.

\subsubsection{Optimization Stability}
\begin{theorem}[Low-frequency Gradient Dominance]
For the \(l\)-th layer weight matrix \(\mathbf{W}^{(l)} = \mathbf{\Lambda}^{(l)}\mathbf{B}^{(l)}\), where \(\mathbf{\Lambda}^{(l)} \in \mathbb{R}^{d \times M}\) and \(\mathbf{B}^{(l)} \in \mathbb{R}^{M \times d}\) with \(M \ll d\), there exists a set of basis matrices \(\{\mathbf{B}^{(l)}\}\) such that for frequencies \(k_1 > k_2 > 0\) and any \(\epsilon \geq 0\):
\begin{equation}
    \left|\frac{\partial \mathbb{L}(k_1)}{\partial \lambda_{ij}^{(l)}} / \frac{\partial \mathbb{L}(k_2)}{\partial \lambda_{ij}^{(l)}}\right| \geq \max_{k}\left|\frac{\partial \mathbb{L}(k_1)}{\partial w_{ik}^{(l)}} / \frac{\partial \mathbb{L}(k_2)}{\partial w_{ik}^{(l)}}\right| - \epsilon.
\end{equation}
\label{theorem: low}
\end{theorem}
The detailed proof of Theorem~\ref{theorem: low} is provided in Appendix~\ref{sec:proof}, following the theoretical framework established in on~\cite{shi2024improved}. This theorem establishes that, during optimization, the gradients associated with low-frequency components consistently dominate those from high-frequency components. As a consequence, the network inherently emphasizes the most influential low-frequency features, thereby preserving its overall representational capacity.

Furthermore, by reparameterizing the weight matrix \(\mathbf{W}^{(l)}\) as the product \(\mathbf{\Lambda}^{(l)} \mathbf{B}^{(l)}\), we effectively reduce the number of trainable parameters from \(d \times d\) to \(d \times M\) (with \(M \ll d\)). This reparameterization not only decreases computational complexity and memory consumption but also facilitates more stable optimization. The preservation of the dominant low-frequency gradient ratio helps mitigate the adverse effects of high-frequency noise, which is often responsible for gradient instability, such as explosion or vanishing gradients. As a result, our approach promotes smoother convergence and improved overall training stability.

\section{Experiments}

\subsection{Experiment settings}

\paragraph{Benchmark}
To rigorously evaluate the effectiveness of LFR-PINO, we conduct experiments on four representative PDEs that cover a broad spectrum of differential operators and physical phenomena:
\begin{itemize}
\item \textbf{Anti-derivative Equation}: A fundamental linear operator that tests the ability to capture integral relationships ($5,000$ training samples);

\item \textbf{Advection Equation}: A first-order hyperbolic PDE with spatially varying coefficients, representing transport phenomena ($1,000$ samples);

\item \textbf{Burgers Equation}: A nonlinear PDE exhibiting both diffusive and convective behavior, crucial for fluid dynamics applications ($500$ diverse initial conditions);

\item \textbf{Diffusion-Reaction System}: A coupled PDE system on a $100\times100$ grid demonstrating complex spatiotemporal dynamics ($1,000$ source-solution pairs).
\end{itemize}
For all cases, we generate the parameter spaces using a Gaussian Random Field (GRF) with zero mean and exponential quadratic kernel:
\begin{equation}
k_l(x_1, x_2) = \exp\left(-\frac{\|x_1-x_2\|^2}{2l^2}\right),
\end{equation}
where the length scale parameter $l$ controls the spatial correlation and smoothness of the sampled functions. We set $l=0.2$ to ensure sufficient complexity while maintaining physical relevance. This sampling strategy ensures our training data encompasses a rich variety of physical scenarios while maintaining mathematical consistency.
Comprehensive problem formulations, boundary conditions, and numerical settings are detailed in Appendix~\ref{app:settings}.

\paragraph{Baselines}
Our baseline selection strategically covers all three physics-informed neural operator paradigms identified in Section 3.1, enabling comprehensive comparison of spectral representation strategies:
\begin{itemize}
\item \textbf{PI-DeepONet}~\cite{wang_learning_2021}: Represents the \textit{dynamic coefficients} category. While the original method does not include fine-tuning, we implement an extended version with fine-tuning capabilities for fair comparison with other methods.
\item \textbf{MAD}~\cite{MADHuang2021}: Exemplifies the \textit{dynamic basis} category. We adopt the MAD-LM variant for fine-tuning, which has demonstrated superior performance in the original work~\cite{MADHuang2021}.
\item \textbf{Hyper-PINNs}~\cite{belbute-peres_hyperpinn_nodate}: Represents the most flexible \textit{dynamic basis and coefficients} category, using a single hypernetwork to generate all parameters of the main network.
\end{itemize}

\paragraph{Implementation Details}
For fair comparison, all methods are implemented using fully connected neural networks with consistent architectures. All experiments are conducted on a single NVIDIA V100 GPU to ensure consistent performance measurements. 
We use GELU activation functions throughout and employ the Adam optimizer for both pre-training and fine-tuning stages. The networks are initialized using truncated normal distribution to ensure stable training.  Detailed configurations for each benchmark problem and baseline method are provided in Appendix~\ref{app:settings}.

\paragraph{Evaluation Metrics}
We evaluate the performance using the relative $L_2$ error:
\begin{equation}
    \text{Error} = \frac{\|u_{\text{pred}} - u_{\text{true}}\|_{L_2}}{\|u_{\text{true}}\|_{L_2}} 
\end{equation}
where $u_{\text{true}}$ represents the reference solution obtained from numerical simulation and $u_{\text{pred}}$ denotes the neural network prediction. This metric is widely used in PDE-related tasks as it provides a normalized measure of solution accuracy independent of the solution magnitude.

\subsection{Pre-training Performance: Universal PDE Solver}

The pre-training results presented in Table~\ref{tab:accuracy-pre-train} highlight significant performance disparities among the different method categories. Both Hyper-PINNs and our LFR-PINO, which utilize dynamic basis and coefficients, demonstrate superior accuracy over PI-DeepONet and MAD, particularly in complex PDEs such as the advection equation and diffusion-reaction system. This underscores the advantage of concurrently learning both basis functions and coefficients to capture a wider range of PDE characteristics. Notably, LFR-PINO consistently outperforms all baselines across all test cases, achieving state-of-the-art accuracy with an error reduction of 4.6\%-68.7\% across various problems.

In terms of computational efficiency and accuracy, our method achieves an optimal balance among existing approaches. While PI-DeepONet and MAD exhibit lower memory usage due to their fixed-basis or coefficient designs, this comes at the cost of reduced solution accuracy. Detailed parametric comparisons are available in Appendix~\ref{app:settings}. As illustrated in Figure~\ref{fig:memory-comparison}, LFR-PINO significantly reduces memory consumption compared to Hyper-PINNs while maintaining excellent accuracy, thanks to our frequency domain reduction strategy. This strategy effectively preserves critical spectral features while eliminating redundant parameters.

\begin{table}[htbp]
  \centering
  \caption{Accuracy of LFR-PINO and baselines in pre-training stage}
  \resizebox{\textwidth}{!}{
  \begin{tabular}{lcccc}
    \toprule
    \textbf{Methods} & \textit{Anti-derivative} & \textit{Advection} & \textit{Burgers} & \textit{Diffusion} \\
    \midrule
    \textbf{PI-DeepONet}~\cite{wang_learning_2021} & 0.00382 & 0.04968 & 0.04543 & 0.08229 \\
    \textbf{MAD}~\cite{MADHuang2021} & 0.02150 & 0.03361 & 0.15861 & 0.17927 \\
    \textbf{Hyper-PINNs}~\cite{belbute-peres_hyperpinn_nodate} & 0.00486 & 0.01982 & 0.04447 & 0.06562 \\
    \rowcolor{gray!20}
    \textbf{LFR-PINO(Ours)} & \textbf{0.00336} & \textbf{0.00621} & \textbf{0.03935} & \textbf{0.03921} \\
    \bottomrule
  \end{tabular}
  }
  \label{tab:accuracy-pre-train}
\end{table}

\begin{figure}[htbp]
    \centering    \subfloat{\includegraphics[width=0.6\textwidth]{ 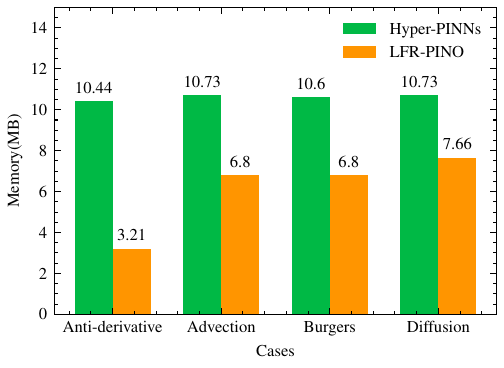}} \hfill
    \caption{Memory comparison between LFR-PINO and HyperPINNs}
    \label{fig:memory-comparison}
\end{figure}

\subsection{Fine-tuning Performance: Adaptation to New Equation}

\begin{figure}[htbp]
    \centering
    \subfloat[Anti-derivative]{\includegraphics[width=0.48\textwidth]{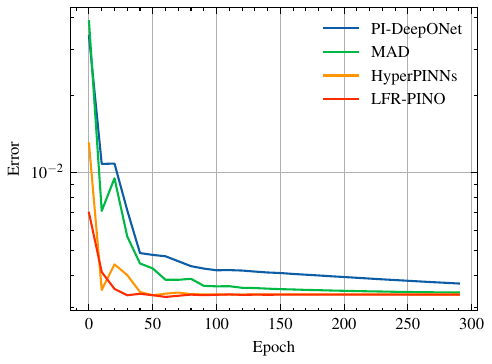}} \hfill 
    \subfloat[Diffusion]{\includegraphics[width=0.48\textwidth]{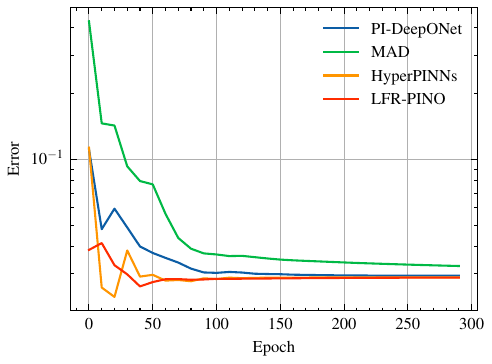}} \hfill 
    \caption{Fine-tuning convergence curves of LFR-PINO and baselines for Anti-derivative and Diffusion cases}
    \label{fig:fine-tune-convergence}
\end{figure}

To validate the practical value of our pre-trained model, we assess its adaptation capability on previously unseen PDEs through one-shot fine-tuning. Table~\ref{tab:accuracy-fine} provides a comparison of performance and computational cost, highlighting the advantages of our approach. During the fine-tuning phase, where a total of 300 epochs are trained, LFR-PINO achieves superior accuracy compared to published baselines. The prediction results and error distributions of LFR-PINO on the test cases after fine-tuning are detailed in Appendix~\ref{app:mapping}. Furthermore, as illustrated in Figure~\ref{fig:fine-tune-convergence}, LFR-PINO converges faster than other methods.

From an inference perspective, our method exhibits advantages in both accuracy and efficiency. This positions LFR-PINO as a highly effective solution for real-time applications, ensuring precise and swift predictions across various PDE scenarios.

\begin{table}[htbp]
  \centering
  \caption{Accuracy of LFR-PINO and baselines in fine-tuning stage}
  \resizebox{\textwidth}{!}{
  \begin{tabular}{lcccc}
    \toprule
    \textbf{Methods} & \textit{Anti-derivative} & \textit{Advection} & \textit{Burgers} & \textit{Diffusion} \\
    \midrule
    \textbf{PI-DeepONet}~\cite{wang_learning_2021} & 0.00371 & 0.04198 & 0.02758 & 0.02938 \\
    \textbf{MAD}~\cite{MADHuang2021} & 0.00343 & 0.03197 & 0.03048 & 0.03243 \\
    \textbf{Hyper-PINNs}~\cite{belbute-peres_hyperpinn_nodate} & 0.00343 & 0.01110 & 0.02188 & 0.02887 \\
    \rowcolor{gray!20}
    \textbf{LFR-PINO(Ours)} & \textbf{0.00326} & \textbf{0.00899} & \textbf{0.01892} & \textbf{0.02818} \\
    \bottomrule
  \end{tabular}
  }
  \label{tab:accuracy-fine}
\end{table}

\subsection{Model Analysis}

\subsubsection{Impact of Frequency-Domain Reduction}
To isolate the effect of frequency-domain reduction, we compare our LFR-PINO with a layered Hyper-PINNs implementation that shares identical network architecture but operates in the full parameter space. This controlled comparison allows us to directly evaluate the impact of our frequency-domain reduction strategy.

\begin{figure}[htbp]
    \centering
    \subfloat[Layered Hyper-PINNs-Layer2]{\includegraphics[width=0.32\textwidth]{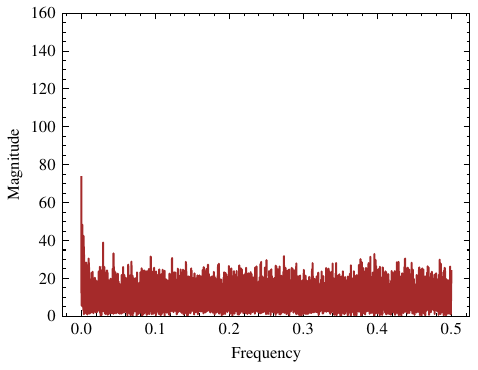}} \hfill
    \subfloat[Layered Hyper-PINNs-Layer4]{\includegraphics[width=0.32\textwidth]{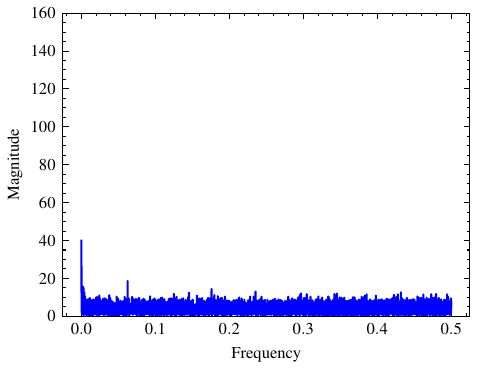}} \hfill
    \subfloat[Layered Hyper-PINNs-Layer6]{\includegraphics[width=0.32\textwidth]{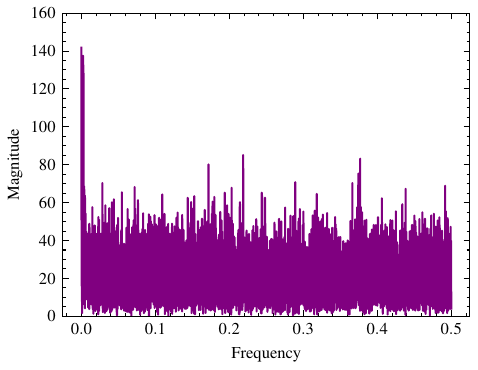}} \hfill \\
    \subfloat[LFR-PINO-Layer2]{\includegraphics[width=0.32\textwidth]{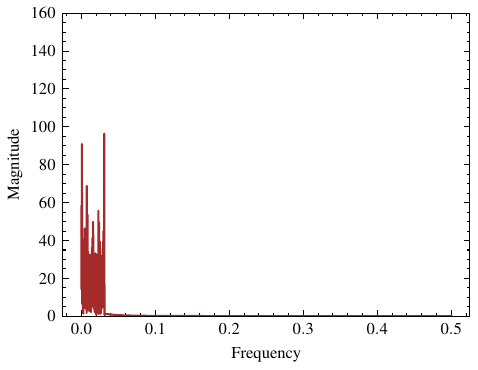}} \hfill
    \subfloat[LFR-PINO-Layer4]{\includegraphics[width=0.32\textwidth]{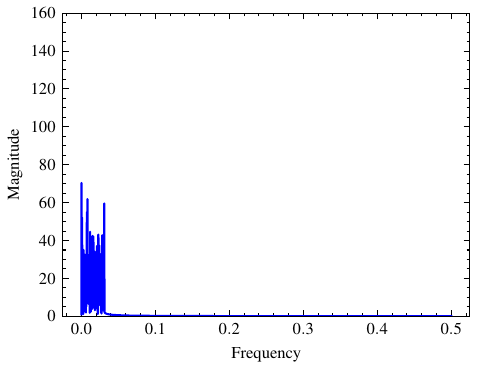}} \hfill
    \subfloat[LFR-PINO-Layer6]{\includegraphics[width=0.32\textwidth]{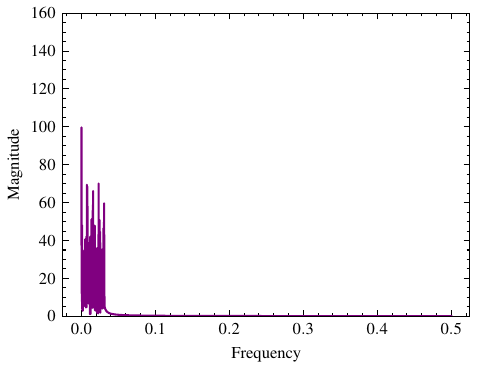}} \hfill
    \caption{Spectral analysis of weight distributions across network layers}
    \label{fig:weight_spec}
\end{figure}

\begin{figure}[htbp]
    \centering
    \subfloat[Layered Hyper-PINNs-Layer2]{\includegraphics[width=0.32\textwidth]{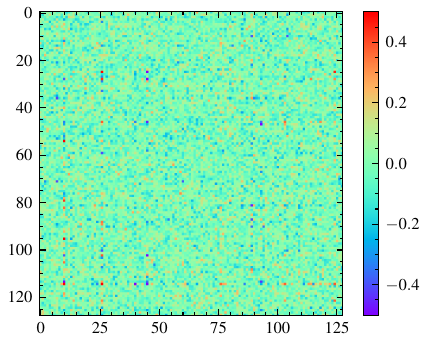}} \hfill
    \subfloat[Layered Hyper-PINNs-Layer4]{\includegraphics[width=0.32\textwidth]{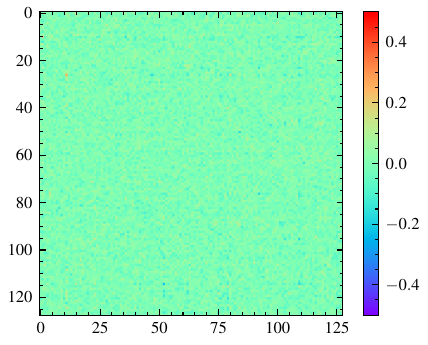}} \hfill
    \subfloat[Layered Hyper-PINNs-Layer6]{\includegraphics[width=0.32\textwidth]{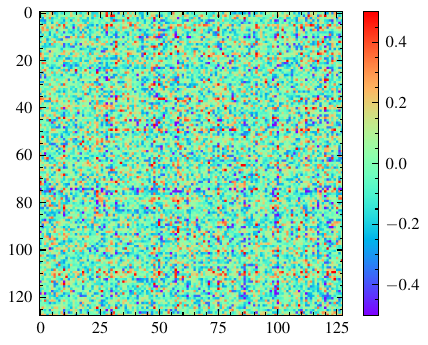}} \hfill \\
    \subfloat[LFR-PINO-Layer2]{\includegraphics[width=0.32\textwidth]{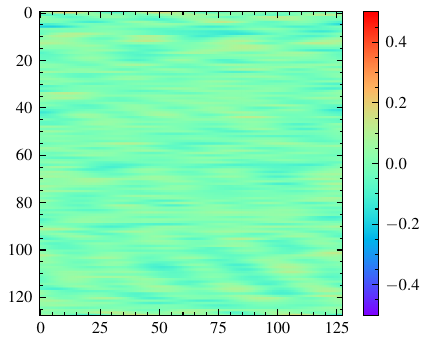}} \hfill
    \subfloat[LFR-PINO-Layer4]{\includegraphics[width=0.32\textwidth]{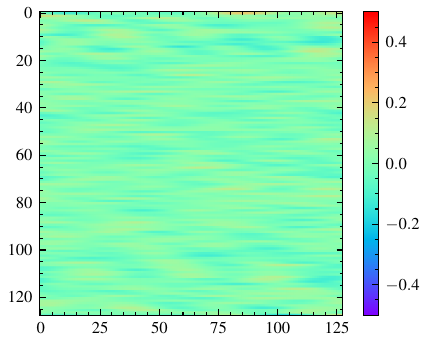}} \hfill
    \subfloat[LFR-PINO-Layer6]{\includegraphics[width=0.32\textwidth]{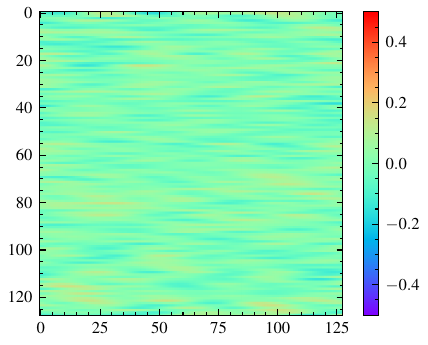}} \hfill
    \caption{Weight distributions across network layers}
    \label{fig:weight_show}
\end{figure}

Figure~\ref{fig:weight_spec} reveals the fundamental spectral differences between these approaches. The layered Hyper-PINNs exhibits uniform modal distributions across all frequency bands, indicating full-spectrum parameter mapping. In contrast, LFR-PINO concentrates its modal distribution in the low-frequency range, demonstrating the effectiveness of our frequency-domain reduction. This spectral concentration is reflected in the weight distributions shown in Figure~\ref{fig:weight_show}, where LFR-PINO displays notably sparser patterns, especially in intermediate hidden layers (2nd and 6th layers), compared to the widespread distributions in layered Hyper-PINNs.

To quantitatively assess the frequency-specific approximation capabilities, we follow Xu et al.~\cite{xu2018Fourier} and analyze the relative difference $\Delta_k$ between target and predicted signals at each frequency $k$:
\begin{equation}
\begin{aligned}
\Delta_k = \frac{\left|\mathcal{F}_D[g](k) - \mathcal{F}_D[f_\theta](k)\right|}{\left|\mathcal{F}_D[g](k)\right|},
\end{aligned}
\label{eq:fourier_error}
\end{equation}
where $\mathcal{F}_D$ denotes the discrete Fourier transform, $g$ is the target solution, and $f_\theta$ is the network prediction.

\begin{figure}[htbp]
    \centering
    \subfloat[Layered Hyper-PINNs]{\includegraphics[width=0.48\textwidth]{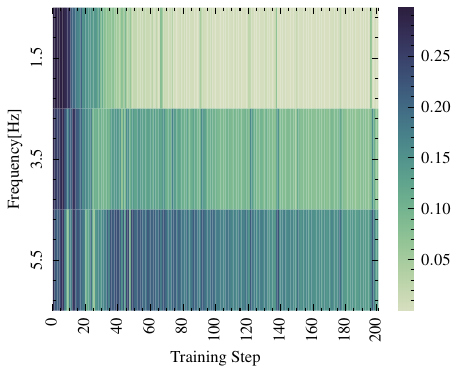}} \hfill 
    \subfloat[LFR-PINO]{\includegraphics[width=0.48\textwidth]{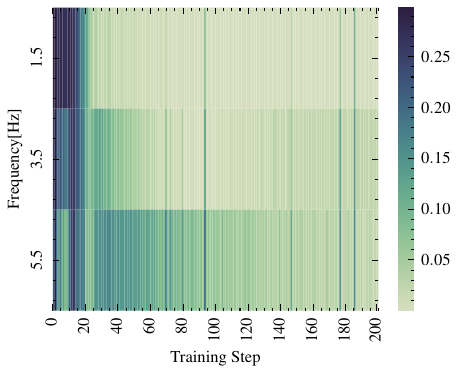}} \hfill 
    \caption{Evolution of frequency-specific approximation error during anti-derivative pre-training (x-axis: training steps, y-axis: frequency components, colormap: relative error)}
    \label{fig:freq-error}
\end{figure}

Figure~\ref{fig:freq-error} compares the evolution of frequency-specific errors during training. LFR-PINO demonstrates two key advantages: faster suppression of high-frequency errors in early training stages and better accuracy in low-frequency components. While our frequency reduction introduces minor spectral oscillations compared to layered Hyper-PINNs, this trade-off enables significant parameter efficiency while preserving essential solution features.

\subsubsection{Influence of Layered Hypernetworks}
\begin{figure}[htbp]
    \centering
    \subfloat[Pre-training]{\includegraphics[width=0.48\textwidth]{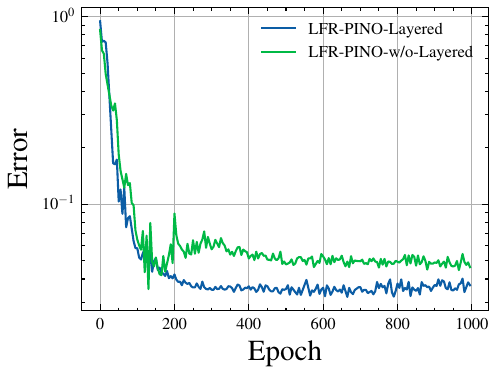}} \hfill 
    \subfloat[Fine-tuning]{\includegraphics[width=0.48\textwidth]{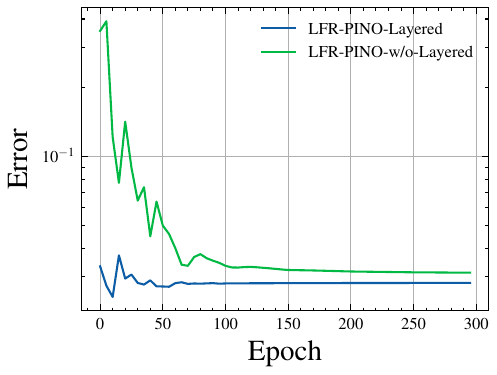}} \hfill 
    \caption{Convergence of LFR-PINO with and without Layered Method on Diffusion Case during pre-training and fine-tuning}
    \label{fig:hypernetwork-version}
\end{figure}

We analyze the impact of hypernetwork architecture by comparing single versus layered designs in the diffusion case. Figure~\ref{fig:hypernetwork-version} shows that our layered approach, which employs multiple specialized hypernetworks, achieves higher accuracy than the single-hypernetwork baseline. This improvement stems from each hypernetwork's ability to focus on optimizing specific layers of the main network, leading to better parameter generation and improved model generalization.
The performance gain, however, comes with increased computational overhead. The layered architecture requires more training epoch compared to the single-hypernetwork design.

\subsection{Ablation Studies}

\subsubsection{Frequency-Domain Reduction Ratio}
\begin{figure}[htpb]
    \centering
    \begin{minipage}[t]{0.48\textwidth}
        \centering
        \includegraphics[width=\textwidth]{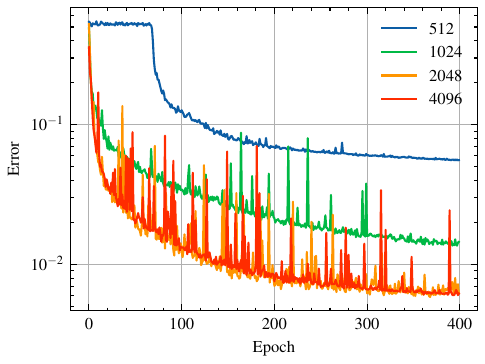}
        \caption{Training convergence under different Fourier modes}
        \label{fig:dim_pretrain}
    \end{minipage}
    \hspace{5pt}
    \begin{minipage}[t]{0.48\textwidth}
        \centering
        \includegraphics[width=\textwidth]{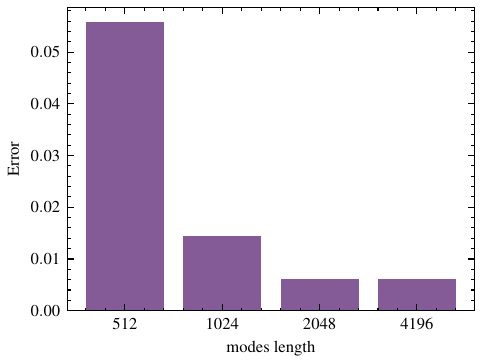}
        \caption{Final accuracy under different Fourier modes}
        \label{fig:result_pre}
    \end{minipage}
\end{figure}

The reduction ratio in frequency domain significantly impacts both performance and efficiency. Using the advection equation as a test case, we investigate different reduction ratios by varying the Fourier modes from 4096 to 512 (original hidden layer dimension: 16512). Figure~\ref{fig:dim_pretrain} shows the convergence histories under different settings, while Figure~\ref{fig:result_pre} presents the final accuracy results.

Higher reduction ratios preserve more frequency components, leading to faster convergence and better accuracy but requiring more memory. Notably, even aggressive reduction (1024 modes) achieves acceptable accuracy, demonstrating our method's ability to capture essential features with minimal parameters. This suggests that most weight information can be effectively encoded in low-frequency components, enabling significant computational savings without severe performance degradation.

\subsubsection{Activation Function}
The choice of activation function significantly influences the spectral properties of neural networks, particularly in our frequency-domain reduction framework. We analyze four common activation functions (GELU, Sigmoid, Sine, and Tanh) by examining their effects on both signal transformation and weight distribution.

\begin{figure}
    \centering
    \subfloat[Standard sine wave]{\includegraphics[width=0.5\textwidth]{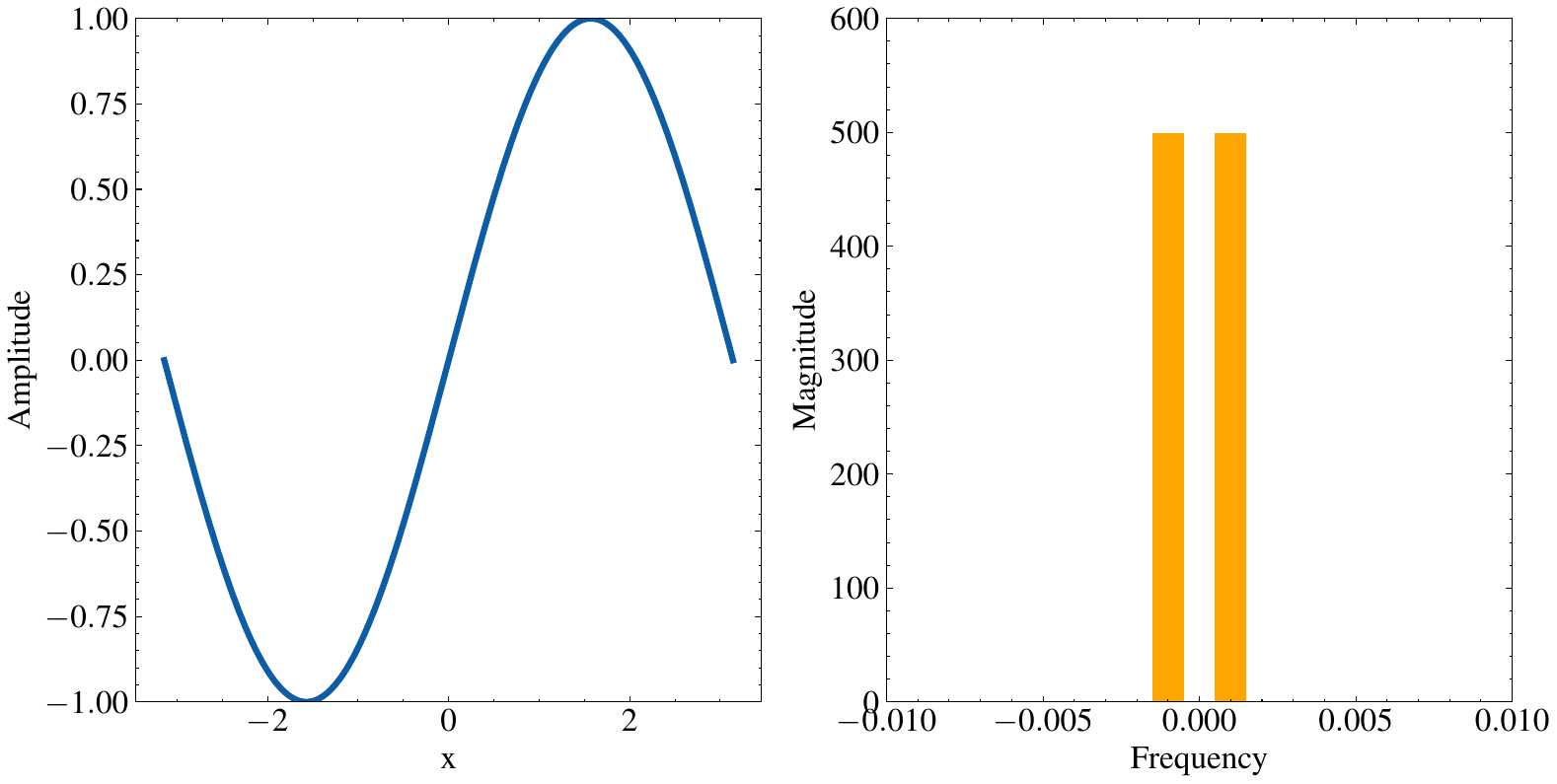}} \hfill \\
    \subfloat[GELU]{\includegraphics[width=0.5\textwidth]{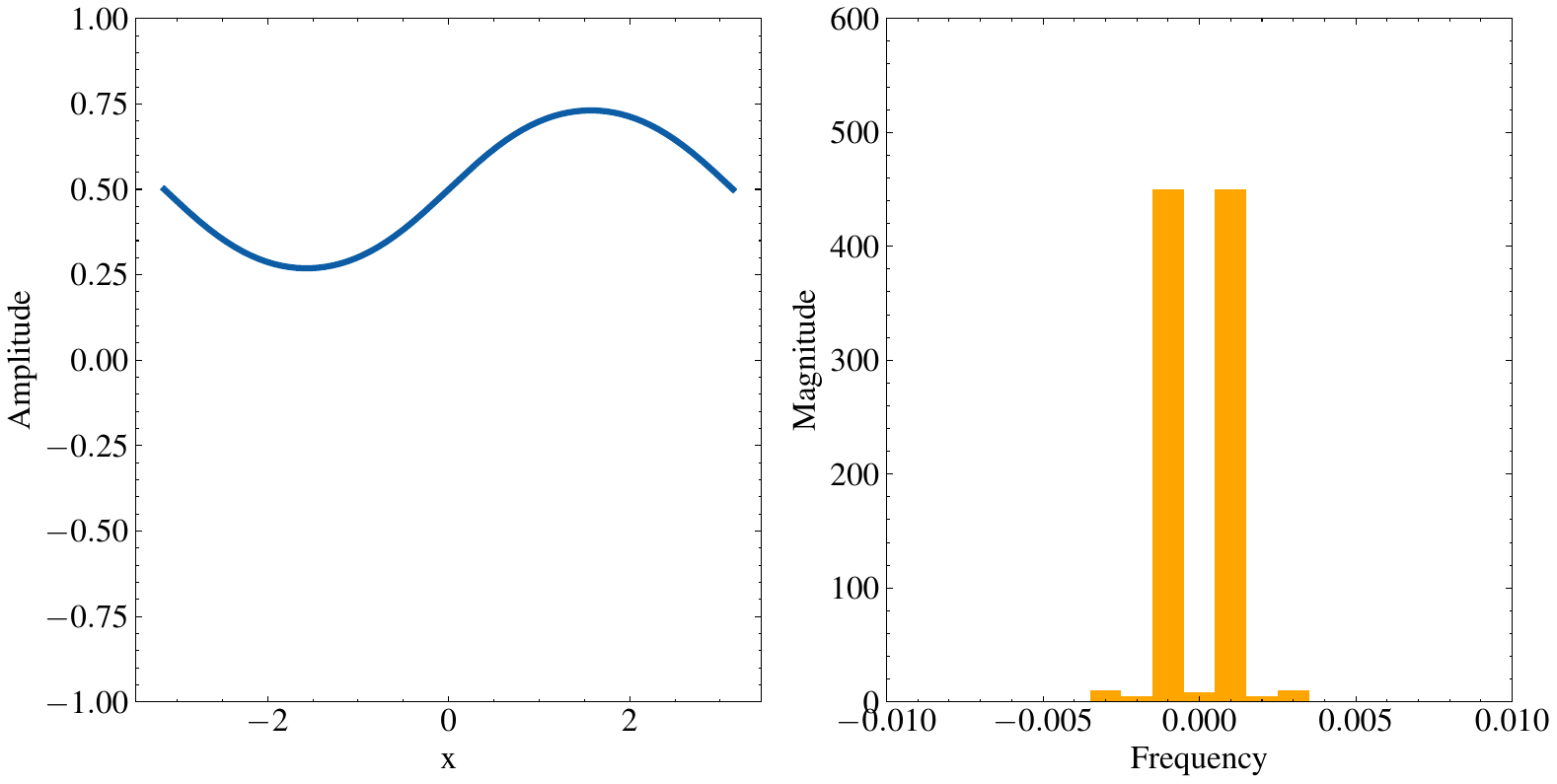}} \hfill 
    \subfloat[Tanh]{\includegraphics[width=0.5\textwidth]{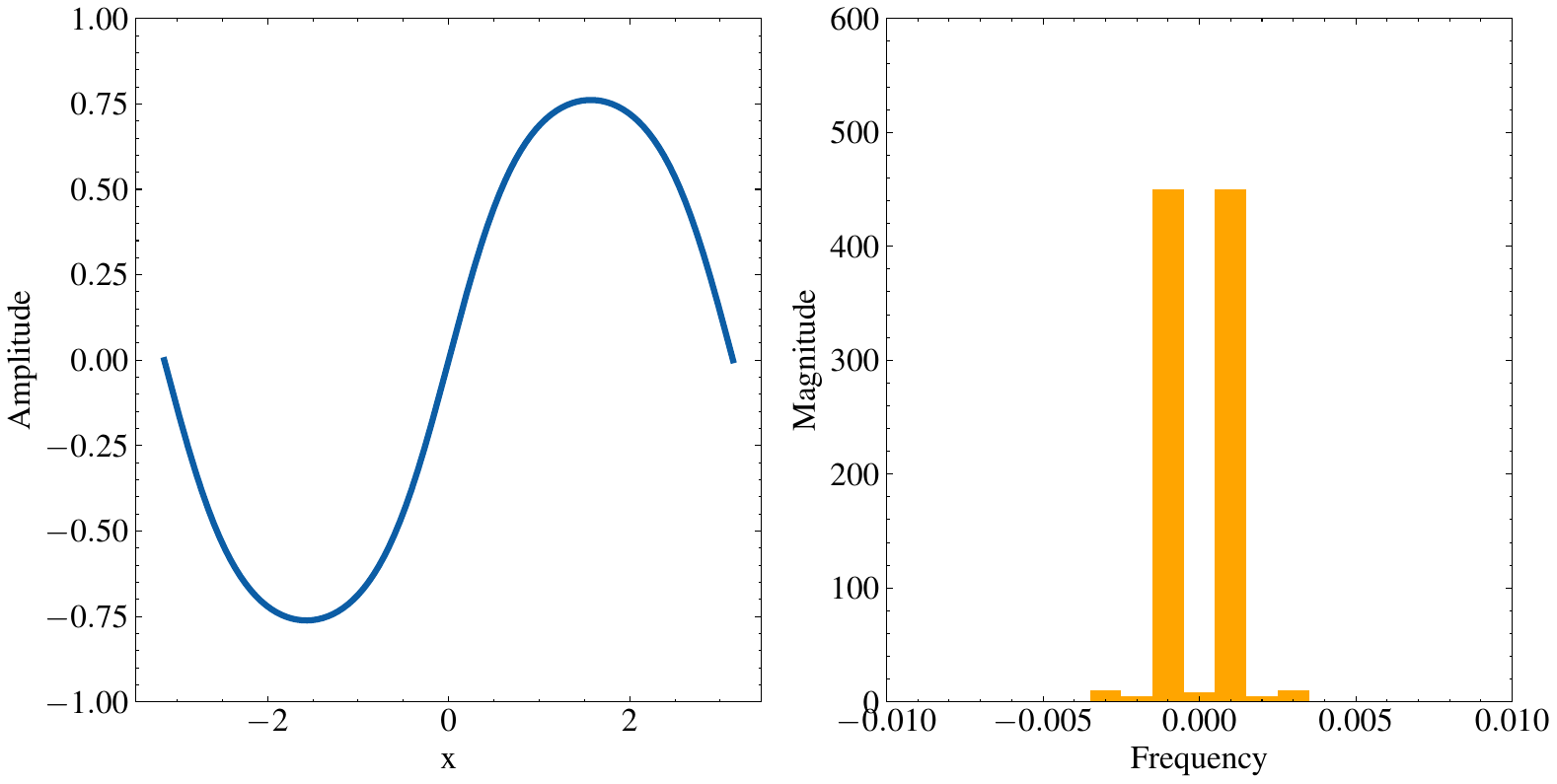}} \hfill \\
    \subfloat[Sine]{\includegraphics[width=0.5\textwidth]{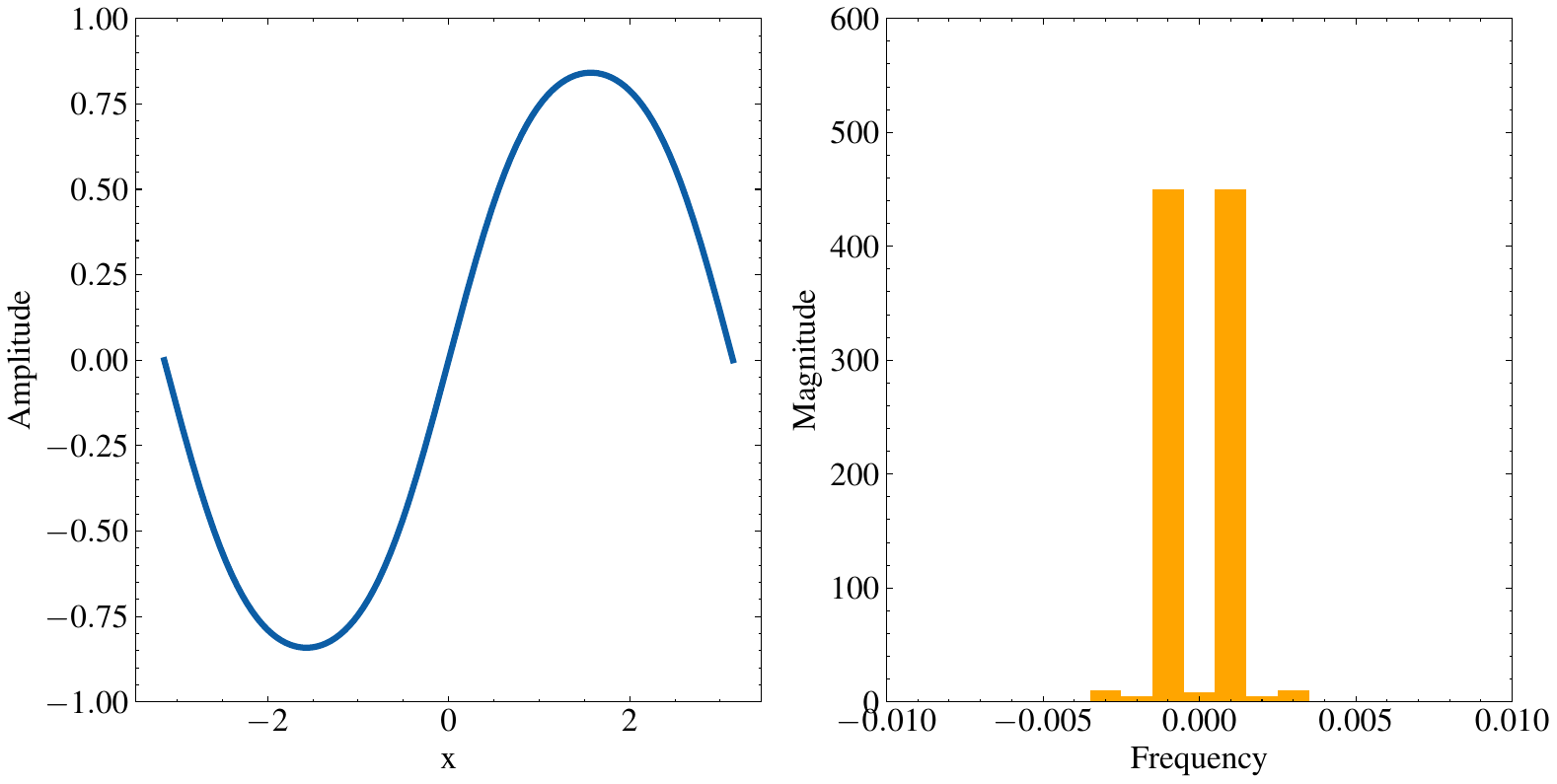}} \hfill
    \subfloat[Sigmoid]{\includegraphics[width=0.5\textwidth]{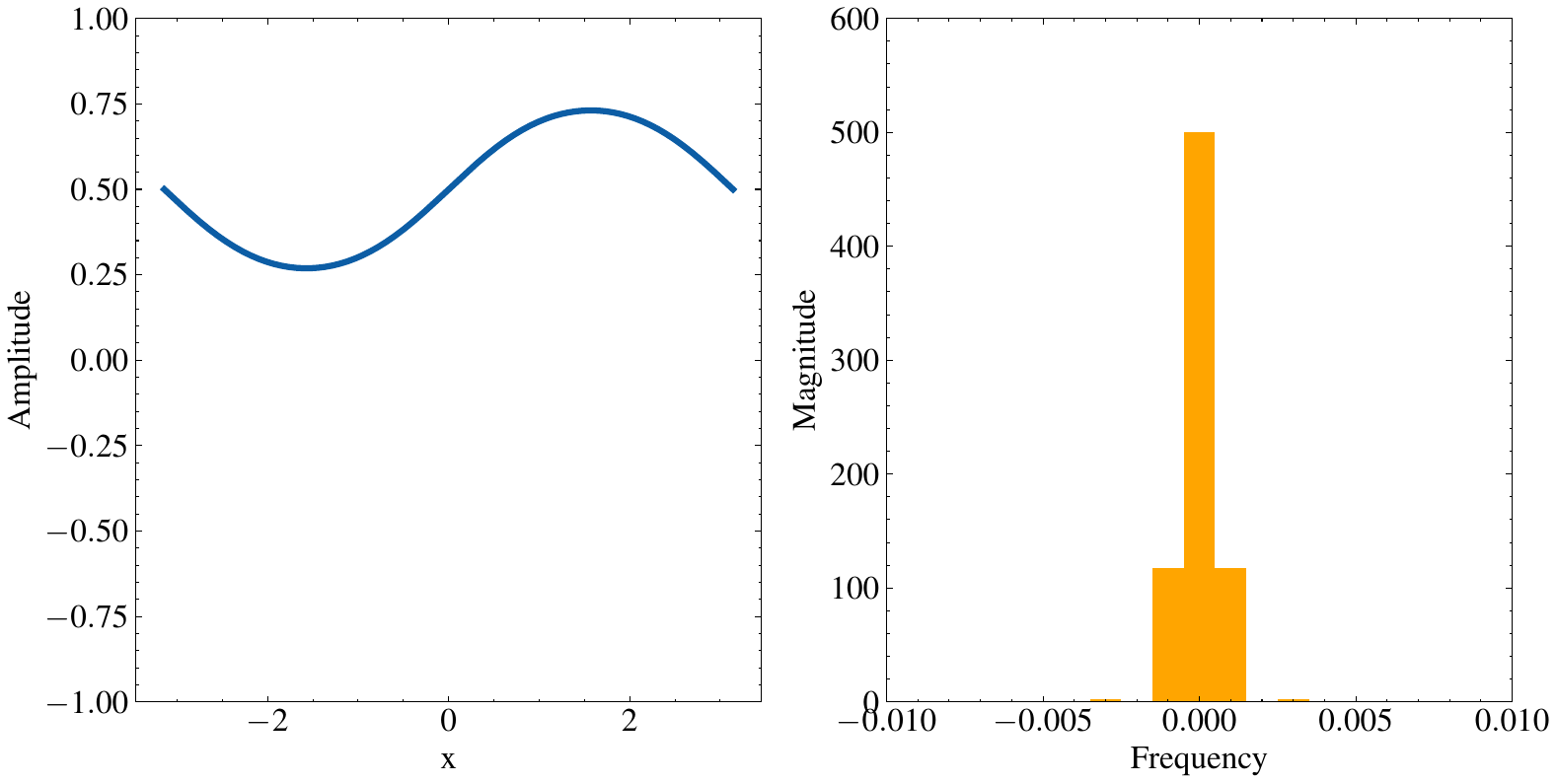}} \hfill 
    \caption{Comparative Analysis of Activation Function Impact on Sine Wave }
    \label{fig:sine-act}
\end{figure}

Figure~\ref{fig:sine-act} demonstrates how each activation function transforms a standard sine wave input. GELU, Sine, and Tanh preserve the main frequency components while introducing minimal high-frequency noise. In contrast, Sigmoid generates more high-frequency components, potentially affecting the network's spectral properties.

\begin{figure}[htbp]
    \centering   \subfloat{\includegraphics[width=0.98\textwidth]{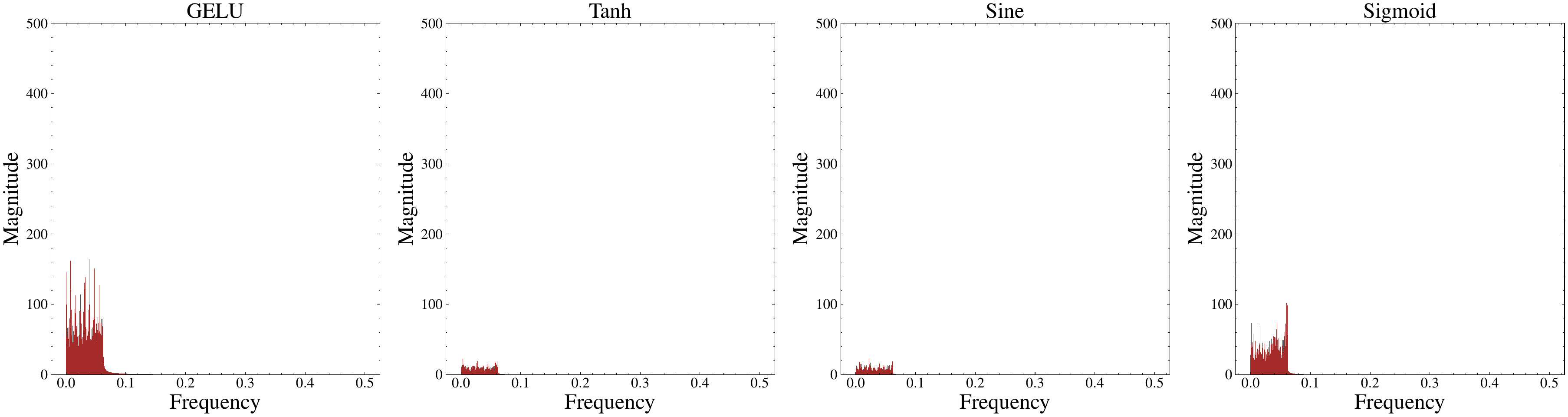}} \hfill
    \caption{Comparative Spectral Weight Distribution of Main Neural Networks under Activation Functions }
    \label{fig:activation-FF}
\end{figure}

This spectral behavior directly impacts model performance, as evidenced by the weight distributions and accuracy results. Figure~\ref{fig:activation-FF} shows the Fourier coefficients of network weights (fourth layer) under different activation functions. Sigmoid exhibits larger high-frequency coefficients, indicating increased sensitivity to spectral noise. GELU, Sine, and Tanh maintain more concentrated spectral distributions, aligning better with our frequency-domain reduction strategy.

The performance impact is quantified in Table~\ref{tab:activation-error}, where GELU achieves the best accuracy (0.0062 relative error) by balancing expressiveness and high-frequency noise suppression. While Sigmoid shows competitive performance (0.0089 error), its tendency to introduce high-frequency components makes it less suitable for our frequency-domain reduction approach.

\begin{table}[htbp]
  \centering
  \caption{Relative $L_2$ error under different activation functions}
  \begin{tabular}{ccccc}
    \toprule
    Activation Function & GELU & Tanh & Sin & Sigmoid \\
    \midrule
    Error & \textbf{0.0062} & 0.0116 & 0.0373 & 0.0089 \\
    \bottomrule
  \end{tabular}
  \label{tab:activation-error}
\end{table}

\section{Conclusions}
We present LFR-PINO, a novel physics-informed neural operator that combines frequency-domain reduction with layered hypernetworks for solving parametric PDEs. Our method achieves state-of-the-art performance through two key innovations: (1) a layered hypernetwork architecture that enables specialized parameter generation for each network layer, and (2) a frequency-domain reduction strategy that significantly reduces parameter count while preserving essential solution features.

Comprehensive experiments on four representative PDE problems demonstrate that LFR-PINO outperforms existing approaches in both accuracy and efficiency. In the pre-training stage, our method achieves 22.8\%-68.7\% error reduction compared to baseline methods while reducing memory usage by up to 69.3\%. The frequency-domain reduction proves particularly effective, allowing us to compress the parameter space to one-fourth of its original size without significant performance degradation. Our ablation studies further reveal that the choice of activation function critically impacts spectral properties, with GELU providing the optimal balance between expressiveness and high-frequency noise suppression.

Looking forward, several promising research directions emerge from this work. The strong performance of frequency-domain reduction suggests potential applications in other deep learning architectures beyond PDE solving. The interaction between parameter space and weight space in the frequency domain warrants further theoretical investigation, particularly in the context of optimization dynamics and generalization bounds. Additionally, our mesh-free and unsupervised approach could be extended to more complex scenarios involving irregular geometries and multi-physics coupling in practical aerospace problems, especially when combined with advanced techniques such as adaptive loss functions~\cite{xiang2022self} or gradient-enhanced optimization~\cite{yu2022gradient}.

\vspace{1cm}
\textbf{Data availability statement}
Data will be made available on request.

\section*{Acknowledgments}
This work is supported by the Natural Science Foundation of China (Nos. U23A2069, 92052203) and the Startup Fund for Young Faculty at SJTU (SFYF at SJTU). This work was partially supported by ZJU Kunpeng\&Ascend Center of Excellence.

\bibliographystyle{tfq}
\bibliography{reference.bib}

\begin{thebibliography}{10}
\newcommand{\printfirst}[2]{#1}
\newcommand{\switchargs}[2]{#2#1}
\providecommand{\url}[1]{\normalfont{#1}}
\providecommand{\urlprefix}{Available at }

\bibitem{li2022machine}
J. Li, X. Du, and J.R. Martins, \emph{Machine learning in aerodynamic shape optimization}, Progress in Aerospace Sciences 134 (2022), p. 100849.

\bibitem{han2018solving}
J. Han, A. Jentzen, and W. E, \emph{Solving high-dimensional partial differential equations using deep learning}, Proceedings of the National Academy of Sciences 115 (2018), pp. 8505--8510.

\bibitem{ZIENKIEWICZ20131}
O. Zienkiewicz, R. Taylor, and J. Zhu, \emph{Chapter 1 - the standard discrete system and origins of the finite element method}, in \emph{The Finite Element Method: its Basis and Fundamentals (Seventh Edition)}, O. Zienkiewicz, R. Taylor, and J. Zhu, eds., seventh edition ed., Butterworth-Heinemann, Oxford,  2013, pp. 1--20, \urlprefix\url{https://www.sciencedirect.com/science/article/pii/B9781856176330000010}.

\bibitem{godunov1959finite}
S.K. Godunov and I. Bohachevsky, \emph{Finite difference method for numerical computation of discontinuous solutions of the equations of fluid dynamics}, Matemati{\v{c}}eskij sbornik 47 (1959), pp. 271--306.

\bibitem{eymard2000finite}
R. Eymard, T. Gallou{\"e}t, and R. Herbin, \emph{Finite volume methods}, Handbook of numerical analysis 7 (2000), pp. 713--1018.

\bibitem{lu2021learning}
L. Lu, P. Jin, G. Pang, Z. Zhang, and G.E. Karniadakis, \emph{Learning nonlinear operators via deeponet based on the universal approximation theorem of operators}, Nature machine intelligence 3 (2021), pp. 218--229.

\bibitem{li2020fourier}
Z. Li, N. Kovachki, K. Azizzadenesheli, B. Liu, K. Bhattacharya, A. Stuart, and A. Anandkumar, \emph{Fourier neural operator for parametric partial differential equations}  (2021).

\bibitem{wang_learning_2021}
S. Wang, H. Wang, and P. Perdikaris, \emph{Learning the solution operator of parametric partial differential equations with physics-informed {DeepOnets}}. \urlprefix\url{http://arxiv.org/abs/2103.10974}.

\bibitem{MADHuang2021}
X. {Huang}, Z. {Ye}, H. {Liu}, B. {Shi}, Z. {Wang}, K. {Yang}, Y. {Li}, B. {Weng}, M. {Wang}, H. {Chu}, F. {Yu}, B. {Hua}, L. {Chen}, and B. {Dong}, \emph{{Meta-Auto-Decoder for Solving Parametric Partial Differential Equations}}, arXiv e-prints  (2021), arXiv:2111.08823.

\bibitem{belbute-peres_hyperpinn_nodate}
F.d.A. Belbute-Peres, F. Sha, and Y.f. Chen, \emph{{HyperPINN}: Learning parameterized differential equations with physics-informed hypernetworks} .

\bibitem{xu2019training}
J. Xu, Y. Zhang, and Y. Xiao, \emph{Training behavior of deep neural network in frequency domain} (2019).

\bibitem{xu2018Fourier}
J. Xu, \emph{Understanding training and generalization in deep learning by fourier analysis}, CoRR abs/1808.04295 (2018), \urlprefix\url{http://arxiv.org/abs/1808.04295}.

\bibitem{lecun2015deep}
Y. LeCun, Y. Bengio, and G. Hinton, \emph{Deep learning}, nature 521 (2015), pp. 436--444.

\bibitem{RAISSI2019686}
M. Raissi, P. Perdikaris, and G. Karniadakis, \emph{Physics-informed neural networks: A deep learning framework for solving forward and inverse problems involving nonlinear partial differential equations}, Journal of Computational Physics 378 (2019), pp. 686--707, \urlprefix\url{https://www.sciencedirect.com/science/article/pii/S0021999118307125}.

\bibitem{sirignano2018dgm}
J. Sirignano and K. Spiliopoulos, \emph{Dgm: A deep learning algorithm for solving partial differential equations}, Journal of computational physics 375 (2018), pp. 1339--1364.

\bibitem{yu2018deep}
B. Yu, \emph{et~al.}, \emph{The deep ritz method: a deep learning-based numerical algorithm for solving variational problems}, Communications in Mathematics and Statistics 6 (2018), pp. 1--12.

\bibitem{rao2023encoding}
C. Rao, P. Ren, Q. Wang, O. Buyukozturk, H. Sun, and Y. Liu, \emph{Encoding physics to learn reaction--diffusion processes}, Nature Machine Intelligence 5 (2023), pp. 765--779.

\bibitem{raissi2020}
M. Raissi, A. Yazdani, and G.E. Karniadakis, \emph{Hidden fluid mechanics: Learning velocity and pressure fields from flow visualizations}, Science 367 (2020), pp. 1026--1030.

\bibitem{xu2021explore}
H. Xu, W. Zhang, and Y. Wang, \emph{Explore missing flow dynamics by physics-informed deep learning: The parameterized governing systems}, Physics of Fluids 33 (2021).

\bibitem{long2018pde}
Z. Long, Y. Lu, X. Ma, and B. Dong, \emph{Pde-net: Learning pdes from data}, in \emph{International conference on machine learning}, PMLR. 2018, pp. 3208--3216.

\bibitem{li2020neural}
Z. Li, N. Kovachki, K. Azizzadenesheli, B. Liu, K. Bhattacharya, A. Stuart, and A. Anandkumar, \emph{Neural operator: Graph kernel network for partial differential equations}, arXiv preprint arXiv:2003.03485  (2020).

\bibitem{deng2023prediction}
Z. Deng, J. Wang, H. Liu, H. Xie, B. Li, M. Zhang, T. Jia, Y. Zhang, Z. Wang, and B. Dong, \emph{Prediction of transonic flow over supercritical airfoils using geometric-encoding and deep-learning strategies}, Physics of Fluids 35 (2023).

\bibitem{penwarden_metalearning_2023}
M. Penwarden, S. Zhe, A. Narayan, and R.M. Kirby, \emph{A metalearning approach for physics-informed neural networks ({PINNs}): Application to parameterized {PDEs}} 477, p. 111912, \urlprefix\url{https://www.sciencedirect.com/science/article/pii/S0021999123000074}.

\bibitem{liu_novel_2022}
X. Liu, X. Zhang, W. Peng, W. Zhou, and W. Yao, \emph{A novel meta-learning initialization method for physics-informed neural networks} 34, pp. 14511--14534, \urlprefix\url{https://link.springer.com/10.1007/s00521-022-07294-2}.

\bibitem{guo__metapinns_2024}
Y. Guo, X. Cao, J. Song, and H. Leng, \emph{{MetaPINNs}: Predicting soliton and rogue wave of nonlinear {PDEs} via the improved physics-informed neural networks based on meta-learned optimization} 33, p. 020203, \urlprefix\url{https://iopscience.iop.org/article/10.1088/1674-1056/ad0bf4}.

\bibitem{zou2023hydra}
Z. Zou and G.E. Karniadakis, \emph{L-hydra: Multi-head physics-informed neural networks}, arXiv preprint arXiv:2301.02152  (2023).

\bibitem{ha2016hypernetworks}
D. Ha, A. Dai, and Q.V. Le, \emph{Hypernetworks}, arXiv preprint arXiv:1609.09106  (2016).

\bibitem{shi2024improved}
K. Shi, X. Zhou, and S. Gu, \emph{Improved Implicit Neural Representation with Fourier Reparameterized Training}, in \emph{Proceedings of the IEEE/CVF Conference on Computer Vision and Pattern Recognition}. 2024, pp. 25985--25994.

\bibitem{xiang2022self}
Z. Xiang, W. Peng, X. Liu, and W. Yao, \emph{Self-adaptive loss balanced physics-informed neural networks}, Neurocomputing 496 (2022), pp. 11--34.

\bibitem{yu2022gradient}
J. Yu, L. Lu, X. Meng, and G.E. Karniadakis, \emph{Gradient-enhanced physics-informed neural networks for forward and inverse pde problems}, Computer Methods in Applied Mechanics and Engineering 393 (2022), p. 114823.

\bibitem{iserles2009first}
A. Iserles, \emph{A first course in the numerical analysis of differential equations}, 44, Cambridge university press, 2009.

\end{thebibliography}

\appendix
\section{Detailed proof of Theorem 2}\label{sec:proof}

\renewcommand{\thetheorem}{2}  
\begin{theorem}[Low-frequency Gradient Theorem]
For the \(l\)-th layer weight matrix \(\mathbf{W}^{(l)} = \mathbf{\Lambda}^{(l)}\mathbf{B}^{(l)}\), where \(\mathbf{\Lambda}^{(l)} \in \mathbb{R}^{d \times M}\) and \(\mathbf{B}^{(l)} \in \mathbb{R}^{M \times d}\) with \(M \ll d\), there exists a set of basis matrices \(\{\mathbf{B}^{(l)}\}\) such that for frequencies \(k_1 > k_2 > 0\) and any \(\epsilon \geq 0\):
\begin{equation}
    \left|\frac{\partial \mathbb{L}(k_1)}{\partial \lambda_{ij}^{(l)}} / \frac{\partial \mathbb{L}(k_2)}{\partial \lambda_{ij}^{(l)}}\right| \geq \max_{k}\left|\frac{\partial \mathbb{L}(k_1)}{\partial w_{ik}^{(l)}} / \frac{\partial \mathbb{L}(k_2)}{\partial w_{ik}^{(l)}}\right| - \epsilon.
\end{equation}
\end{theorem}
\renewcommand{\thetheorem}{\arabic{theorem}}  

\begin{proof}
Before the detailed proof, simply denote: \(\mathbb{L}_{\lambda_{ij}^{(l)}}(k_1) = \frac{\partial \mathbb{L}(k_1)}{\partial \lambda_{ij}^{(l)}}\), \(\mathbb{L}_{w_{ij}^{(l)}}(k_1) = \frac{\partial \mathbb{L}(k_1)}{\partial w_{ij}^{(l)}}\).

First, the weight reparameterization for \(\mathbf{W}^{(l)}\) is expressed as follows:
\[
\mathbf{W}^{(l)} = \mathbf{\Lambda}^{(l)} \mathbf{B}^{(l)},
\]
by the matrix multiplication, for any \(w_{ij}^{(l)} \in \mathbf{W}^{(l)}\), the follow equation holds true:
\[
w_{ij}^{(l)} = \begin{bmatrix} \lambda_{i1}^{(l)}, \lambda_{i2}^{(l)}, \ldots, \lambda_{iM}^{(l)} \end{bmatrix} \begin{bmatrix} b_{1j}^{(l)} \\ b_{2j}^{(l)} \\ \vdots \\ b_{Mj}^{(l)} \end{bmatrix},
\]
where \(B^{(l)}(i,j) = b_{ij}^{(l)}\). Regarding \(w_{i1}^{(l)}, \ldots, w_{id}^{(l)}\) as the latent variables related with \(\lambda_{ij}^{(l)}\), for all \(\lambda_{ij}^{(l)} \in \mathbf{\Lambda}^{(l)}\), using the chain rule, we have the following relationships:
\[
\mathbb{L}_{\lambda_{ij}^{(l)}}(k) = \sum_{t=1}^d b_{jt}^{(l)} \mathbb{L}_{w_{it}^{(l)}}(k).
\]
Second, given two frequencies \(k_1 > k_2 > 0\), for the \(i\)-th row of \(\mathbf{\Lambda}^{(l)}\), we set that:
\[
\tau = \arg\max_j \left\{ \left| \frac{\mathbb{L}_{w_{ij}^{(l)}}(k_1)}{\mathbb{L}_{w_{ij}^{(l)}}(k_2)} \right| \right\}.
\]
Further, considering the elements of \(\mathbf{B}^{(l)}\), for \(j = 1, \ldots, M\), we make \(|b_{jt}^{(l)}| < \alpha\) for \(t \neq \tau\) and \(b_{j\tau}^{(l)} = 1\). \(\alpha\) is a positive upper bound. Then, according to equation 5, for the fixed \(i\), for \(j = 1, \ldots, M\), we have:
\[
\mathbb{L}_{\lambda_{ij}^{(l)}}(k_1) = \mathbb{L}_{w_{i\tau}^{(l)}}(k_1) + \sum_{t \neq \tau} b_{jt}^{(l)} \mathbb{L}_{w_{it}^{(l)}}(k_1),
\]
\[
\mathbb{L}_{\lambda_{ij}^{(l)}}(k_2) = \mathbb{L}_{w_{i\tau}^{(l)}}(k_2) + \sum_{t \neq \tau} b_{jt}^{(l)} \mathbb{L}_{w_{it}^{(l)}}(k_2).
\]
We denote \(G_1\) and \(G_2\) as \(\sum_{t \neq \tau} |\mathbb{L}_{w_{it}^{(l)}}(k_1)|\) and \(\sum_{t \neq \tau} |\mathbb{L}_{w_{it}^{(l)}}(k_2)|\), respectively. Without loss of generality, for any
\[
0 \leq \epsilon \leq \frac{|\mathbb{L}_{w_{i\tau}^{(l)}}(k_1)|}{|\mathbb{L}_{w_{i\tau}^{(l)}}(k_2)|},
\]
set
\[
\alpha \leq \min\left\{ \frac{|\mathbb{L}_{w_{i\tau}^{(l)}}(k_2)| \epsilon}{G_1 + G_2 |\mathbb{L}_{w_{i\tau}^{(l)}}(k_1) / \mathbb{L}_{w_{i\tau}^{(l)}}(k_2)| - G_2 \epsilon}, \frac{|\mathbb{L}_{w_{i\tau}^{(l)}}(k_1)|}{G_1} \right\},
\]
then by inequalities involving absolute values, we have:
\[
\left| \frac{\mathbb{L}_{\lambda_{ij}^{(l)}}(k_1)}{\mathbb{L}_{\lambda_{ij}^{(l)}}(k_2)} \right| = \left| \frac{\mathbb{L}_{w_{i\tau}^{(l)}}(k_1) + \sum_{t \neq \tau} b_{jt}^{(l)} \mathbb{L}_{w_{it}^{(l)}}(k_1)}{\mathbb{L}_{w_{i\tau}^{(l)}}(k_2) + \sum_{t \neq \tau} b_{jt}^{(l)} \mathbb{L}_{w_{it}^{(l)}}(k_2)} \right| \geq \frac{|\mathbb{L}_{w_{i\tau}^{(l)}}(k_1)| - \left| \sum_{t \neq \tau} b_{jt}^{(l)} \mathbb{L}_{w_{it}^{(l)}}(k_1) \right|}{|\mathbb{L}_{w_{i\tau}^{(l)}}(k_2)| + \left| \sum_{t \neq \tau} b_{jt}^{(l)} \mathbb{L}_{w_{it}^{(l)}}(k_2) \right|}.
\]
From 9, we have that \(\alpha \leq \frac{|\mathbb{L}_{w_{i\tau}^{(l)}}(k_1)|}{G_1}\). Then:

\[
\left| \sum_{t \neq \tau} b_{jt}^{(l)} \mathbb{L}_{w_{it}^{(l)}}(k_1) \right| \leq \alpha \sum_{t \neq \tau} |\mathbb{L}_{w_{it}^{(l)}}(k_1)| = \alpha G_1 \leq |\mathbb{L}_{w_{i\tau}^{(l)}}(k_1)|,
\]
which means that \(|\mathbb{L}_{w_{i\tau}^{(l)}}(k_1)| - \left| \sum_{t \neq \tau} b_{jt}^{(l)} \mathbb{L}_{w_{it}^{(l)}}(k_1) \right| \geq 0\). Thus, the following inequalities holds true:
\[
\frac{|\mathbb{L}_{w_{i\tau}^{(l)}}(k_1)| - \left| \sum_{t \neq \tau} b_{jt}^{(l)} \mathbb{L}_{w_{it}^{(l)}}(k_1) \right|}{|\mathbb{L}_{w_{i\tau}^{(l)}}(k_2)| + \left| \sum_{t \neq \tau} b_{jt}^{(l)} \mathbb{L}_{w_{it}^{(l)}}(k_2) \right|} \geq \frac{|\mathbb{L}_{w_{i\tau}^{(l)}}(k_1)| - \alpha G_1}{|\mathbb{L}_{w_{i\tau}^{(l)}}(k_2)| + \alpha G_2} \geq 0.
\]
Substituting \(\alpha \leq \frac{|\mathbb{L}_{w_{i\tau}^{(l)}}(k_2)| \epsilon}{G_1 + G_2 |\mathbb{L}_{w_{i\tau}^{(l)}}(k_1) / \mathbb{L}_{w_{i\tau}^{(l)}}(k_2)| - G_2 \epsilon}\) into the above inequality, for the fixed \(i\) and for \(j = 1, \ldots, M\), we have that:
\[
\left| \frac{\mathbb{L}_{\lambda_{ij}^{(l)}}(k_1)}{\mathbb{L}_{\lambda_{ij}^{(l)}}(k_2)} \right| \geq \frac{|\mathbb{L}_{w_{i\tau}^{(l)}}(k_1)| - \alpha G_1}{|\mathbb{L}_{w_{i\tau}^{(l)}}(k_2)| + \alpha G_2} \geq \frac{|\mathbb{L}_{w_{i\tau}^{(l)}}(k_1)| (G_1 + G_2 |\mathbb{L}_{w_{i\tau}^{(l)}}(k_1) / \mathbb{L}_{w_{i\tau}^{(l)}}(k_2)| - G_2 \epsilon) - G_1 |\mathbb{L}_{w_{i\tau}^{(l)}}(k_2)| \epsilon}{|\mathbb{L}_{w_{i\tau}^{(l)}}(k_2)| (G_1 + G_2 |\mathbb{L}_{w_{i\tau}^{(l)}}(k_1) / \mathbb{L}_{w_{i\tau}^{(l)}}(k_2)| - G_2 \epsilon) + G_2 |\mathbb{L}_{w_{i\tau}^{(l)}}(k_2)| \epsilon}
\]
\[
= \frac{G_1 |\mathbb{L}_{w_{i\tau}^{(l)}}(k_1)| + G_2 \frac{|\mathbb{L}_{w_{i\tau}^{(l)}}(k_1)|^2}{|\mathbb{L}_{w_{i\tau}^{(l)}}(k_2)|}}{G_1 |\mathbb{L}_{w_{i\tau}^{(l)}}(k_2)| + G_2 |\mathbb{L}_{w_{i\tau}^{(l)}}(k_1)|} - \epsilon
\]
\[
= \left| \frac{\mathbb{L}_{w_{i\tau}^{(l)}}(k_1)}{\mathbb{L}_{w_{i\tau}^{(l)}}(k_2)} \right| \cdot \frac{G_1 + G_2 \frac{|\mathbb{L}_{w_{i\tau}^{(l)}}(k_1)|}{|\mathbb{L}_{w_{i\tau}^{(l)}}(k_2)|}}{G_1 + G_2 \frac{|\mathbb{L}_{w_{i\tau}^{(l)}}(k_1)|}{|\mathbb{L}_{w_{i\tau}^{(l)}}(k_2)|}} - \epsilon
\]
\[
= \left| \frac{\mathbb{L}_{w_{i\tau}^{(l)}}(k_1)}{\mathbb{L}_{w_{i\tau}^{(l)}}(k_2)} \right| - \epsilon
\]
\[
= \max\left\{ \left| \frac{\mathbb{L}_{w_{i1}^{(l)}}(k_1)}{\mathbb{L}_{w_{i1}^{(l)}}(k_2)} \right|, \ldots, \left| \frac{\mathbb{L}_{w_{id}^{(l)}}(k_1)}{\mathbb{L}_{w_{id}^{(l)}}(k_2)} \right| \right\} - \epsilon.
\]
Thus, the theorem is proved.
\end{proof}

\section{Detailed Experimental Settings and Results}
\label{app:settings}

\subsection{Anti-derivative Equation}
\paragraph{Problem Definition}
\begin{equation}
\begin{cases}
 \frac{ds}{dx} = u(x), & x \in[0,1] \\
 s(0) = 0 
\end{cases}
\end{equation}

\paragraph{Dataset Settings} 
The training dataset consists of 5,000 input-output pairs, where the input $u(x)$ is sampled from a Gaussian Random Field (GRF) with a correlation length of $l=0.2$. The reference solutions are computed using the RK45 method~\cite{iserles2009first}. The test dataset includes 100 input-output pairs, with one additional sample reserved for fine-tuning.

\paragraph{Implementation Details}
Unspecified, all methods use the Adam optimizer with an initial learning rate of 0.0005, a learning rate of 0.0001 during the fine-tuning phase, and a decay of 0.8 at 100-step intervals. Unspecified, all networks contain four hidden layers with a width of 64. The Fourier transformations of LFR-PINO for the input, hidden and output layers are 32 respectively. truncations are 32, 2048 and 16. We train a total of 500 epochs in the pre-training stage and 300 epochs in the fine-tuning stage.

\begin{figure}
    \centering
    \subfloat
    {\includegraphics[width=0.5\textwidth]{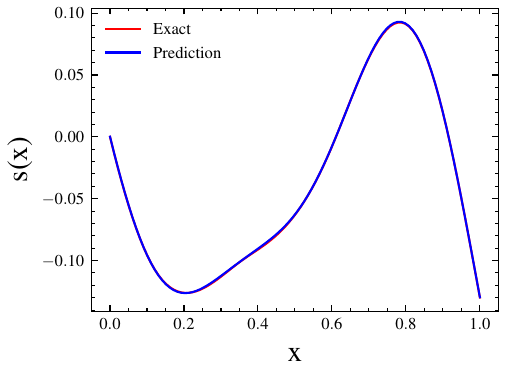}}
    \hfill
    \subfloat
    {\includegraphics[width=0.48\textwidth]{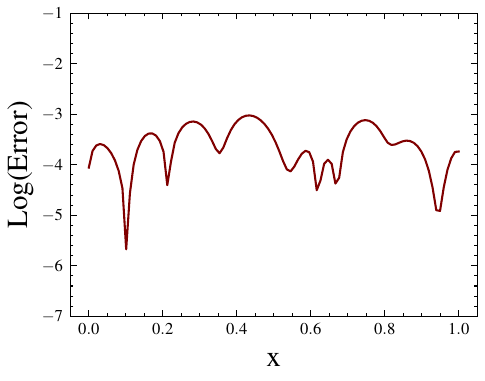}}
    \caption{Analytical solutions and model predictions of LFR-PINO for anti-derivative operator}
    \label{fig:antiderivative2}
\end{figure}

\paragraph{Results}
Figure~\ref{fig:antiderivative2} shows the prediction accuracy of LFR-PINO. The left panel displays the comparison between predicted and analytical solutions, while the right panel shows the pointwise error distribution. The size of the PI-DeepONet and MAD models is  0.13 MB, while the size of the HyperPINNs is 10.44 MB and LFR-PINO is 3.21 MB.

\subsection{Advection Equation}
\paragraph{Problem Definition}
\begin{equation}
\begin{cases}
\frac{\partial s}{\partial t} + a(x)\frac{\partial s}{\partial x} = 0, & (x,t) \in[0,1]^2 \\
s(x,0) = \sin(\pi x) \\
s(0,t) = s(1,t) = \sin(\pi t/2)
\end{cases}
\end{equation}

\paragraph{Dataset Settings}
The training dataset consists of 1,000 coefficient-solution pairs, where the variable coefficient $a(x)$ is sampled from a Gaussian Random Field (GRF) with a correlation length of $l = 0.2$. The test dataset includes one held-out sample.

\paragraph{Implementation Details}
Without special instructions, the network hidden layer widths were all 128, with an initial learning rate of 0.0005 and decayed at a 0.8 multiplicative rate for each of the first 300 epochs of 50 epochs. The Fourier transformations of LFR-PINO for the input, hidden and output layers are 32 respectively. truncations are 32, 2048 and 16. We train a total of 1000 epochs in the pre-training stage and 300 epochs in the fine-tuning stage.

\begin{figure}
    \centering
    \subfloat[Ground Truth]{\includegraphics[width=0.32\textwidth]{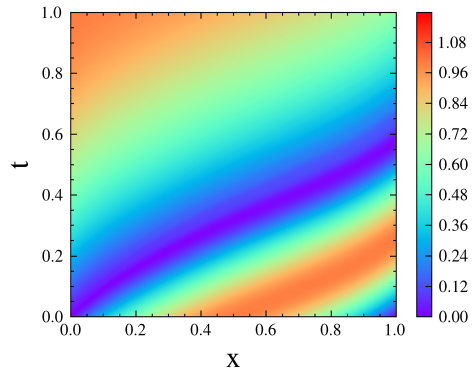}}
    \hfill
    \subfloat[Prediction]{\includegraphics[width=0.32\textwidth]{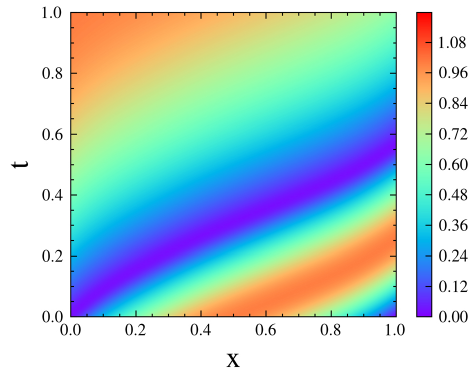}}
    \hfill
    \subfloat[Error]{\includegraphics[width=0.33\textwidth]{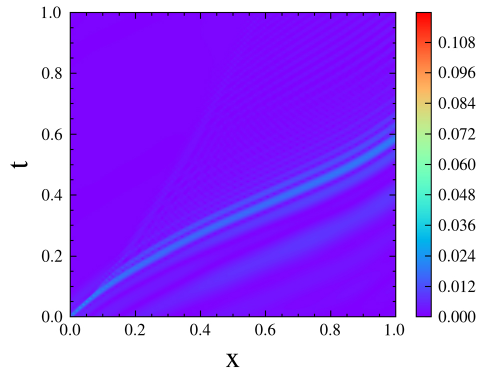}}
    \caption{Analytical solutions and model predictions of LFR-PINO for advection equation}
    \label{fig:ad-visual}
\end{figure}

\paragraph{Results}
As shown in Figure~\ref{fig:ad-visual}, our method accurately captures the advection dynamics. The three panels present (from left to right): exact solution, predicted solution, and absolute error distribution. The size of the PI-DeepONet and MAD models is 0.64 MB, while the size of the HyperPINNs is 10.73 MB MB and LFR-PINO is 6.80 MB.

\subsection{Burgers Equation}
\paragraph{Problem Definition}
\begin{equation}
\begin{cases}
\frac{\partial s}{\partial t} + s\frac{\partial s}{\partial x} = \nu\frac{\partial^2 s}{\partial x^2}, & (x,t) \in[0,1]^2 \\
s(x,0) = u(x) \\
s(0,t) = s(1,t), \quad \frac{\partial s}{\partial x}(0,t) = \frac{\partial s}{\partial x}(1,t)
\end{cases}
\end{equation}

\paragraph{Dataset Settings}
The training dataset consists of 500 initial condition-solution pairs, where the initial profile $u(x)$ is sampled from a Gaussian Random Field (GRF) with a correlation length of $l = 2.5$. The viscosity coefficient is set to $\nu = 0.01$ .

\paragraph{Implementation Details}
The initial learning rate is set to 0.0005, and the decay rate is 0.7 for the first 300 epochs, with 50 epochs allocated for each of the first 300 epochs. The pre-training phase involves the training of 500 epochs, while the fine-tuning phase trains 300 epochs.The Fourier truncation of LFR-PINO for the input, hidden, and output layers is 64, 2048, and 32, respectively.The remaining parameters are identical to those employed in the advection algorithm.

\begin{figure}[htbp]
    \centering
    \subfloat[Ground Truth]{\includegraphics[width=0.32\textwidth]{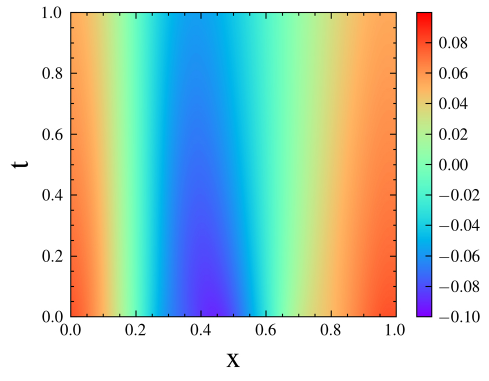}} \hfill
    \subfloat[Predictions]{\includegraphics[width=0.32\textwidth]{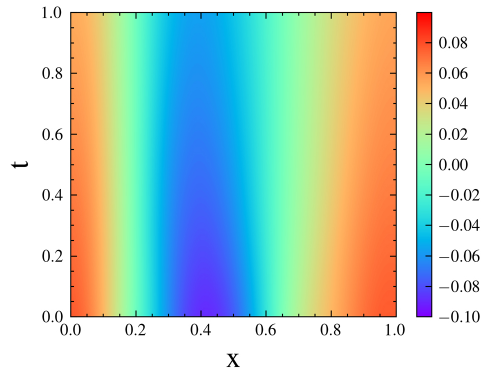}} \hfill
    \subfloat[Error]{\includegraphics[width=0.32\textwidth]{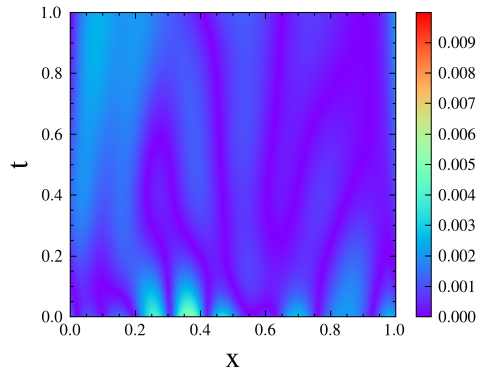}} \\
    \caption{Analytical solutions and model predictions of LFR-PINO for burgers equation}
    \label{fig:burgers-visual}
\end{figure}

\paragraph{Results}
Figure~\ref{fig:burgers-visual} demonstrates the performance on the Burgers equation. The solution field comparison shows excellent agreement between predicted and analytical solutions, with error primarily concentrated in regions of high gradients. The size of the PI-DeepONet and MAD models is 0.16 MB, while the size of the HyperPINNs is 10.60 MB MB and LFR-PINO is 6.79 MB.

\subsection{Diffusion-Reaction System}
\paragraph{Problem Definition}
\begin{equation}
\begin{cases}
\frac{\partial s}{\partial t} = D\frac{\partial^2 s}{\partial x^2} + ks^2 + u(x), & (x,t) \in(0,1]^2 \\
s(0,t) = s(x,0) = 0
\end{cases}
\end{equation}

\paragraph{Dataset Settings}
The training dataset consists of 1,000 source-solution pairs, where the source terms $u(x)$ are sampled from a Gaussian Random Field (GRF) with a correlation length of $l=0.2$. The grid resolution is set to $100×100$ spatial-temporal points, with parameters $D=0.01$ and $k=0.01$.

\paragraph{Implementation Details}
Unless otherwise indicated, all networks have a hidden layer width of 128, an initial learning rate of 0.001, and decay at a multiplicity of 0.5 every 50 epochs for the first 300 epochs. The number of iterations for the pre-training and fine-tuning phases is 1000 and 300 epochs, respectively. The number of truncations of the inputlayer, hidden layers and output layer of the main network by LFR-PINO is 64, 2048, and 32, respectively.

\begin{figure}[htbp]
    \centering
    \subfloat[Ground Truth]{\includegraphics[width=0.32\textwidth]{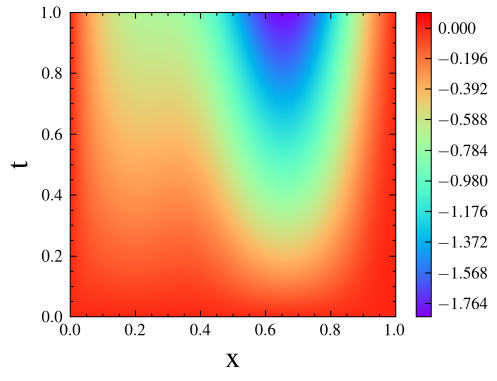}} \hfill
    \subfloat[Predictions]{\includegraphics[width=0.32\textwidth]{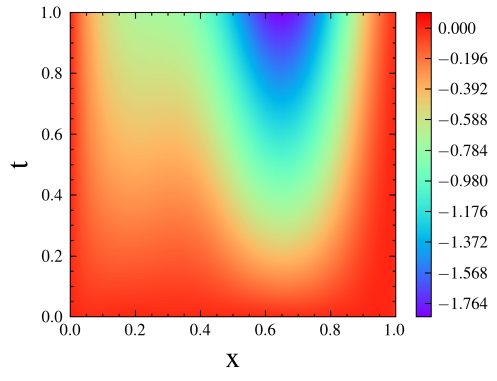}} \hfill
    \subfloat[Error]{\includegraphics[width=0.32\textwidth]{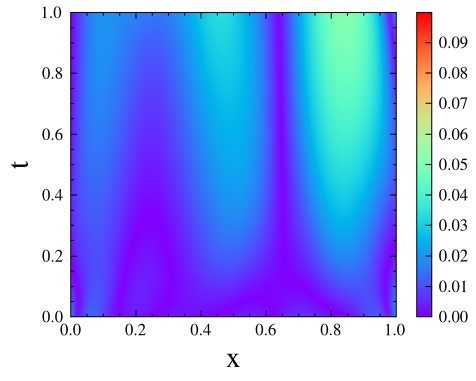}}\\
    \caption{Analytical solutions and model predictions of LFR-PINO for diffusion reaction system}
    \label{fig:diff-visual}
 \end{figure}

\paragraph{Results}
The results for the diffusion-reaction system are presented in Figure~\ref{fig:diff-visual}. The comparison between exact and predicted solutions demonstrates the method's capability in handling coupled nonlinear PDEs, with error analysis showing consistent accuracy across the spatial-temporal domain. The size of the PI-DeepONet and MAD models is 0.77 MB, while the size of the HyperPINNs is 10.73 MB MB and LFR-PINO is 7.66 MB.

\section{Analysis of Parameter-Weight Space Mapping}\label{app:mapping}
To better understand why our frequency-domain reduction strategy works, we analyze the continuity of mapping between parameter space and weight space. This analysis provides insights into how network weights adapt to parameter changes and justifies our focus on low-frequency components.

We use the Burgers equation as a case study, with initial conditions parameterized by a Gaussian function:
\begin{equation}
    s(x, 0)=\exp\left[-\frac{(x-x_0)^2}{2}\right]
\end{equation}
where $x_0$ takes values of 0.4, 0.5, and 2.0, representing small and large parameter variations. The network architecture consists of 5 hidden layers with 50 nodes each, and all models are trained to achieve relative errors below 0.02.

\begin{figure}[htbp]
    \centering
    \subfloat[case1-layer1]{\includegraphics[width=0.32\textwidth]{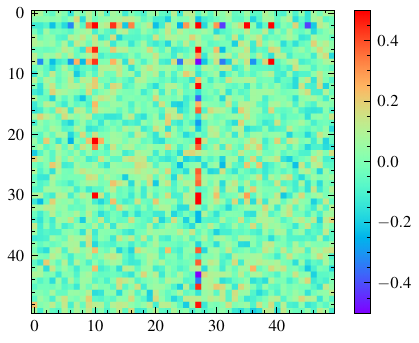}} \hfill
    \subfloat[case2-layer1]{\includegraphics[width=0.32\textwidth]{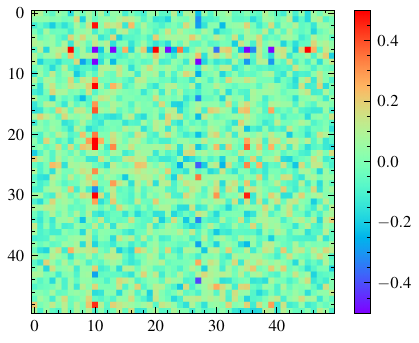}} \hfill
    \subfloat[case3-layer1]{\includegraphics[width=0.32\textwidth]{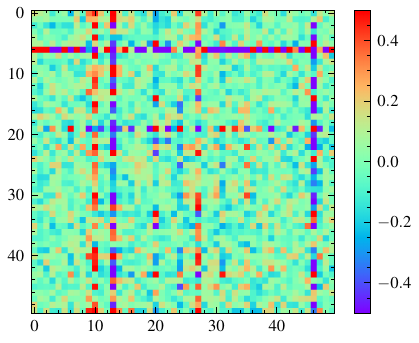}} \\
    \subfloat[case1-layer4]{\includegraphics[width=0.32\textwidth]{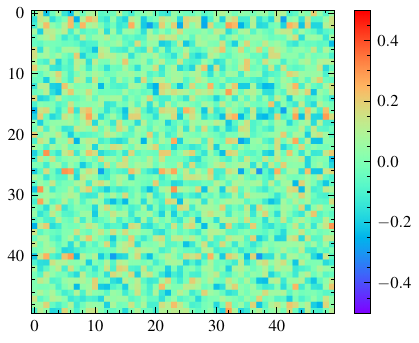}} \hfill
    \subfloat[case2-layer4]{\includegraphics[width=0.32\textwidth]{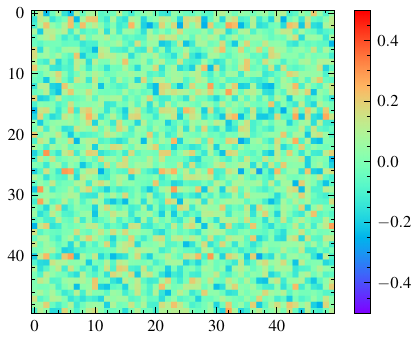}} \hfill
    \subfloat[case3-layer4]{\includegraphics[width=0.32\textwidth]{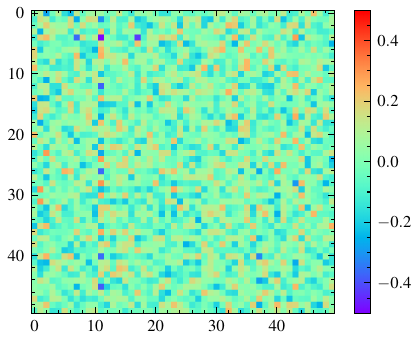}} \\
    \caption{Weight distributions of PINNs}
    \label{fig:visual1}
\end{figure}

Figure~\ref{fig:visual1} reveals two key observations about the weight distributions:
1. Similar parameters lead to similar weight patterns: cases 1 ($x_0=0.4$) and 2 ($x_0=0.5$) show more similar weight distributions compared to case 3 ($x_0=2.0$).
2. Deeper layers exhibit stronger pattern consistency, suggesting that our layered hypernetwork design is well-motivated.

\begin{figure}[htbp]
\centering
\includegraphics[width=0.5\textwidth]{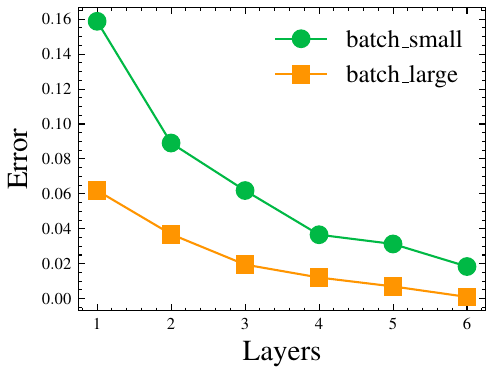}
    \caption{The L1 Paradigm of $\left\|W^{case1}-W^{case2}\right\|$ and $\left\|W^{case2}-W^{case3}\right\|$ in each layer}
    \label{fig:visual2}
\end{figure}

The quantitative analysis in Figure~\ref{fig:visual2} confirms this continuity property:
\begin{equation}
    \|W^{case1}-W^{case2}\| < \|W^{case2}-W^{case3}\| \quad \text{corresponds to} \quad |x_0^{case1}-x_0^{case2}| < |x_0^{case2}-x_0^{case3}|
\end{equation}

This continuous mapping property supports our frequency-domain reduction strategy: if small parameter changes lead to small weight changes, then the essential features of the weight space can be captured by low-frequency components, allowing effective dimensionality reduction without significant loss of expressiveness.

\end{document}